%% file: arxiv.tex
\documentclass[runningheads]{llncs}
\PassOptionsToPackage{dvipsnames}{xcolor}

\usepackage[mobile]{eccv}

\usepackage{eccvabbrv}
\usepackage{array}
\usepackage{multirow}
\usepackage{graphicx}
\usepackage{booktabs}
\newcolumntype{L}[1]{>{\raggedright\arraybackslash}p{#1}}
\usepackage{makecell}
\usepackage{colortbl,xcolor,booktabs}
\usepackage{arydshln}
\usepackage{wrapfig}
\usepackage{caption}

\newcommand{\method}{UMO\xspace}
\newcommand{\myparagraph}[1]{\vspace{0.2cm}\noindent\textbf{#1}}

\usepackage[accsupp]{axessibility}  

\usepackage{hyperref}

\usepackage{orcidlink}
\usepackage{tikz}
\newcommand{\colorball}[1]{%
  \raisebox{-0.2ex}{\textcolor{#1}{\scriptsize$\bullet$}}%
}

\begin{document}

\title{\method: Unified In-Context Learning Unlocks Motion Foundation Model Priors}

\titlerunning{\method}

\author{
Xiaoyan Cong\inst{1*} \and
Zekun Li\inst{1*} \and
Zhiyang Dou\inst{2} \and
Hongyu Li\inst{1} \and
Omid Taheri\inst{4} \and
Chuan Guo\inst{3} \and
Abhay Mittal\inst{3} \and
Sizhe An\inst{3} \and
Taku Komura\inst{5} \and 
Wojciech Matusik\inst{2} \and 
Michael J. Black\inst{4} \and 
Srinath Sridhar\inst{1}}

\authorrunning{Cong et al.}

\institute{$^1$Brown University \quad $^2$Massachusetts Institute of Technology \quad $^3$Meta Reality Lab $^4$Max-Planck Institute for Intelligent Systems \quad \quad $^5$University of Hong Kong}

\maketitle
\renewcommand{\thefootnote}{*}
\footnotetext{Equal contribution.}
\renewcommand{\thefootnote}{\arabic{footnote}}

\begin{figure*}[t]
    \centering
    \includegraphics[width=\textwidth]{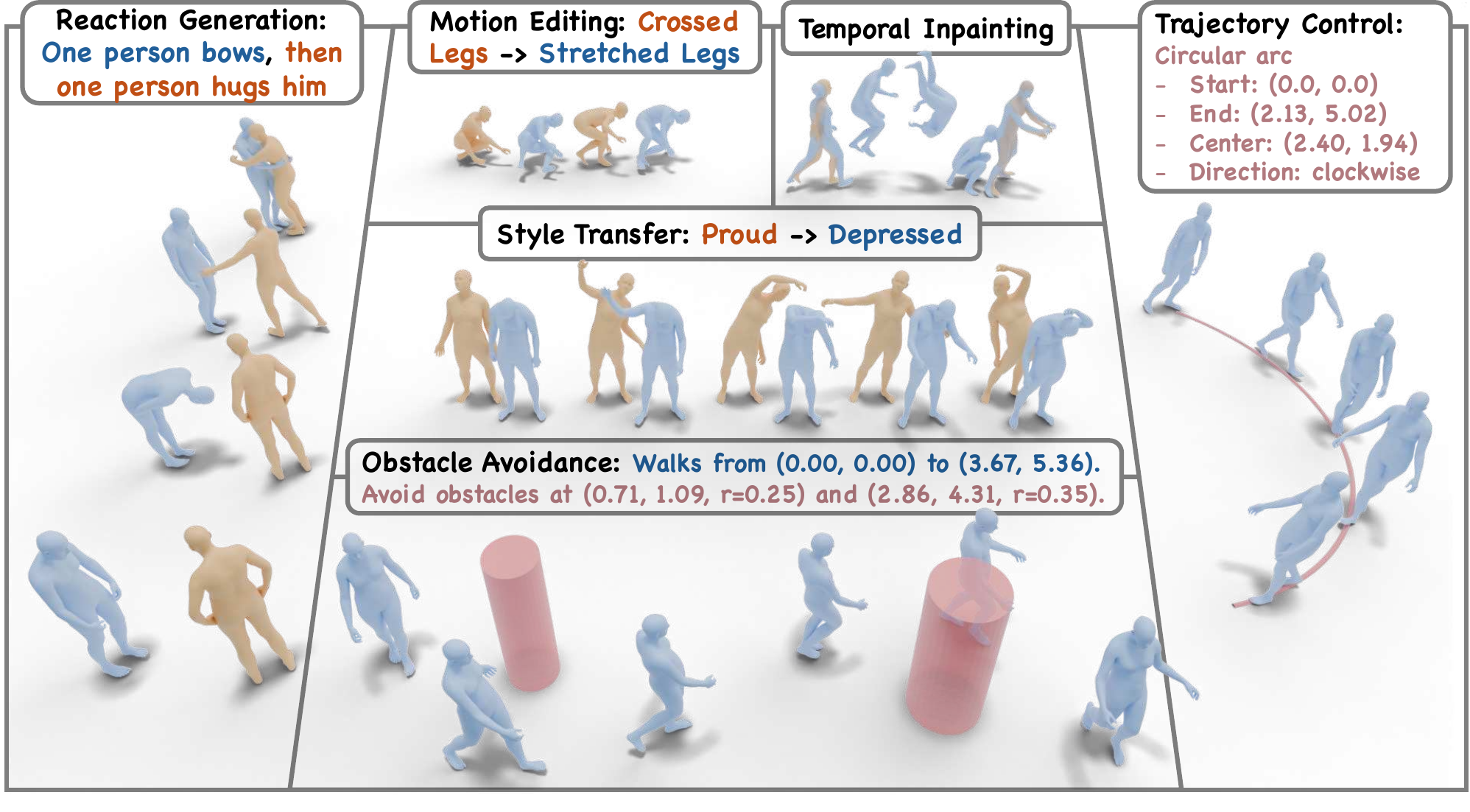}
    \vspace{-0.75cm}
    \caption{\textbf{\method} casts diverse downstream tasks as compositions of our novel atomic per-frame operations, unlocking the generative priors of a pretrained text-to-motion LFM without any architectural modification in a broad spectrum of applications.}
    \vspace{-0.7cm}
    \label{fig:teaser}
\end{figure*}

\input{section/00_abstract}

\input{section/01_introduction}
\input{section/02_related_work}
\input{section/03_method}

\input{section/04_experiment}
\input{section/05_limitation}
\input{section/06_conclusion}

\paragraph{\textbf{Acknowledgements.}}
This research was supported by NSF CAREER grant \#2143576, and a sponsored research award from Meta Inc.

%
%
\bibliographystyle{splncs04}
\bibliography{main}

\clearpage
\input{section/x_supp}

\end{document}

%% file: section/00_abstract.tex
\begin{abstract}
Large-scale foundation models (LFMs) have recently made impressive progress in text-to-motion generation by learning strong generative priors from massive 3D human motion datasets and paired text descriptions. 
However, how to effectively and efficiently leverage such single-purpose motion LFMs, \ie, text-to-motion synthesis, in more diverse cross-modal and in-context motion generation downstream tasks remains largely unclear. 
Prior work typically adapts pretrained generative priors to individual downstream tasks in a task-specific manner. 
In contrast, our goal is to unlock such priors to support a broad spectrum of downstream motion generation tasks within a single unified framework.
To bridge this gap, we present \textbf{\method}, a simple yet general unified formulation that casts diverse downstream tasks into compositions of atomic per-frame operations, enabling in-context adaptation to unlock the generative priors of pretrained DiT-based motion LFMs.
Specifically, \method introduces \textbf{three} \textbf{learnable} \textbf{frame-level} \textbf{meta-operation} \textbf{embeddings} to specify per-frame intent and employs lightweight temporal fusion to inject in-context cues into the pretrained backbone, with negligible runtime overhead compared to the base model.
With this design, \method finetunes the pretrained model, originally limited to text-to-motion generation, to support diverse previously unsupported tasks, including temporal inpainting, text-guided motion editing, text-serialized geometric constraints, and multi-identity reaction generation.
Experiments demonstrate that \method consistently outperforms task-specific and training-free baselines across a wide range of benchmarks, despite using a single unified model. 
Code and models will be publicly available.
Project Page: \href{https://oliver-cong02.github.io/UMO.github.io/}{https://oliver-cong02.github.io/UMO.github.io/}
\keywords{Human Motion Generation \and Motion Foundation Model}
\end{abstract}

%% file: section/01_introduction.tex
\section{Introduction}
\label{sec:introduction}

3D human motion generation~\cite{feng2024motionwavelet, cong2024laserhuman,xu2025mospa,chen2024pay,pinyoanuntapong2025maskcontrol,zhao2026comovi,li2025genmo,liu2025ponimator} has witnessed substantial progress across diverse tasks, yet the field remains fragmented, with each task typically addressed by a specialized architecture, preventing knowledge sharing and cross-task generalization.
In other generative domains~\cite{wanx_ace,wu2025omnigen2,wei2025univideo,cong2025viva,mao2025ace++}, it has been widely demonstrated that unifying diverse tasks under a shared formulation can effectively unlock powerful generative priors from pretrained large foundation models~\cite{flux2024,wan2025wan,kong2024hunyuanvideo,esser2024scaling}.
However, such a unified framework remains largely absent in the motion domain, despite being critical for improving generalization and consolidating scarce per-task data into a unified training framework.
\input{Figure/radar}

Recent efforts in building motion LFMs~\cite{fan2025go, lu2025scamo, li2026llamo, wen2025hy} have significantly scaled up both model capacity and training data, learning general generative priors for human motion.
However, these models primarily focus on basic text-to-motion (T2M) generation, leaving their rich generative priors untapped for broader applications. 
This raises a natural question: \emph{how can we develop a unified and efficient framework that unlocks pretrained motion LFM priors to support diverse unseen downstream tasks?}

We introduce \textbf{\method}, a simple yet effective unified framework that unlocks generative priors of pretrained DiT-based motion LFMs for diverse downstream tasks, as summarized in~\cref{tab:task_formulation}.
Unlike prior task-specific solutions~\cite{shafir2023human,guo2025motionlab,chen2024pay,cohan2024flexible,sawdayee2025dance} that require dedicated architectures or conditioning modules for each task, \method expresses all cross-modal and in-context motion generation tasks under a unified formulation without modifying the pretrained backbone.
Our key insight is that any motion-related task can be decomposed at the frame level into exactly one of three mutually exclusive meta-operations with respect to the source motion context: {it is either preserved (identity), generated without reference motion (context-free generation), or generated based on existing content (context-based generation)}.
Regardless of whether a task involves temporal inpainting, text-guided motion editing, spatial control, or  multi-identity reaction generation, the per-frame intent always reduces to one of these atomic meta-operations. 
Based on this observation, we introduce three learnable frame-level meta-operation embeddings (\texttt{[preserve]}, \texttt{[generate]}, and \texttt{[edit]}) that specify per-frame intent to the generative backbone. 
On the language conditioning side, all task conditions, from semantic descriptions and editing instructions to precise 3D spatial constraints, are expressed as text and encoded by the same pretrained LLM, requiring no task-specific conditioning module.

Specifically, we adopt HY-Motion~\cite{wen2025hy} as our motion LFM backbone, a large-scale DiT-based T2M model that provides robust generative priors.
By augmenting each source motion frame with its corresponding meta-operation embedding and injecting the resulting task-aware sequence into the pretrained backbone via lightweight temporal fusion, our framework extends the T2M-only LFM to naturally cover a broad spectrum of previously unsupported tasks (see~\cref{fig:teaser}), including temporal inpainting~\cite{cohan2024flexible} (\eg, prediction, backcasting, in-betweening, and keyframe control) and text-guided motion editing via either fine-grained instructions~\cite{athanasiou2024motionfix} or high-level style prompts~\cite{jiang2025dynamic}.
Furthermore, \method enables geometric constraints~\cite{wan2024tlcontrol,pinyoanuntapong2025maskcontrol,karunratanakul2024optimizing} in text-format without optimization and multi-identity reaction generation~\cite{xu2024regennet,javed2024intermask,cen2025ready,fan2024freemotion} in a single unified model.
Notably, all these tasks lie beyond the base model's original text-to-motion training scope, yet are handled by a single unified model through two key technical contributions: (1)~frame-level meta-operation embeddings that explicitly specify per-frame generative intent, and (2)~a lightweight and effective temporal fusion mechanism that injects this task-aware conditioning into the DiT backbone, requiring no task-specific modules or architectural changes.
Comprehensive experiments show that the generative prior learned from a \textit{single pretraining objective} (text-to-motion) can be efficiently unlocked to support \textit{various downstream motion generation tasks}, shown in~\cref{fig:radar}.
We also conduct systematic ablations among different conditioning architectures, including temporal fusion~\cite{cong2025viva,mou2025instructx}, sequence concatenation~\cite{tevet2024closd,chen2024taming}, AdaLN~\cite{peebles2023scalable}, and ControlNet~\cite{zhang2023adding}.
Achieving state-of-the-art performance across these tasks with a single unified model demonstrates both the transferability and generalization of motion priors learned from large-scale T2M pretraining, and the effectiveness of our in-context motion conditioning design.

In summary, our contributions are:
\vspace{-0.3cm}
\begin{itemize}
\item \textbf{A unified in-context formulation grounded in three atomic meta-operations.} 
We identify that per-frame intent in any in-context motion task reduces to one of three mutually exclusive operations—preserve, generate, or edit. The corresponding learnable embeddings, combined with unified language conditioning, express diverse downstream tasks without architectural changes to the LFM backbone.
\item \textbf{A lightweight and efficient conditioning design with systematic validation.} 
We propose temporal fusion to inject in-context cues at the input embedding level with only 0.207M additional parameters and negligible runtime overhead, and validate this choice against alternative architectures.
\item \textbf{Evidence of broad prior transferability.} A single unified \method consistently outperforms task-specific baselines across temporal completion, text-guided editing, geometric constraints, and multi-identity reaction generation, all beyond the base model's original training scope.
\end{itemize}

%% file: Figure/radar.tex
\begin{wrapfigure}{r}{0.5\textwidth}
    \centering
    \vspace{-0.2cm}
    \captionsetup{justification=raggedright,singlelinecheck=false}
    \includegraphics[width=0.5\textwidth]{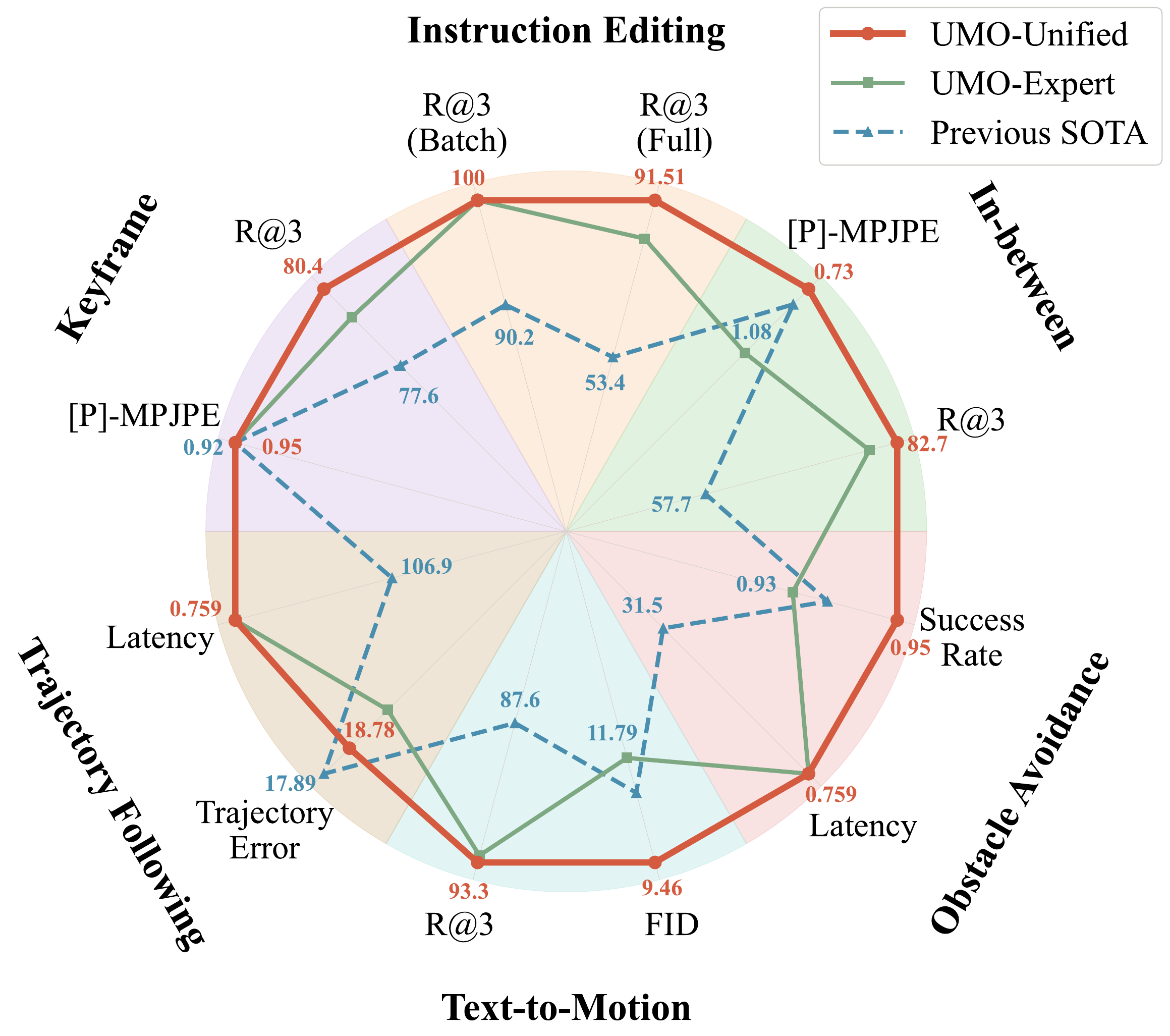}
    \vspace{-0.6cm}
    
    \caption{\method achieves state-of-the-art performance and efficiency across diverse tasks. Latency, Traj.\,Err, $\texttt{[P]}$-MPJPE, and FID are inverted so that the outermost boundary is consistently optimal.}
    \vspace{-1cm}
    \label{fig:radar}
\end{wrapfigure}

%% file: section/02_related_work.tex
\section{Related Work}
\label{sec:related_work}
\subsection{Diverse Motion Generation Downstream Tasks}
Motion generation downstream tasks can be organized around a fundamental distinction: whether the model operates solely from external signals, or must additionally comprehend and respect existing motion context.
\textbf{Cross-modal motion generation}, \ie, $(C_0,C_1,\cdots)\mapsto X$, requires mapping from one or more conditioning modalities into plausible motion sequences. 
Representative modalities include audio~\cite{ng2026sarah,chen2025language,xu2025mospa,ghosh2025duetgen}, 3D environments~\cite{jiang2024scaling,cong2024laserhuman,li2025genhsi,hwang2025scenemi}, spatial constraints~\cite{xie2023omnicontrol,wan2024tlcontrol,pinyoanuntapong2025maskcontrol,karunratanakul2024optimizing}, 3D objects~\cite{xu2024interdreamer,xu2023interdiff,xu2025intermimic,xu2025interact,peng2025hoi}, videos~\cite{lin2025quest,zhao2026comovi,zhang2025egoreact,yu2025hero}, or their combinations. 
\textbf{In-context motion generation}, \ie, $(X,C)\mapsto X$, requires balancing faithfulness to the source motion with responsiveness to the condition, based on overall comprehension.
This category can be further divided along two axes.
\textit{Homogeneous} tasks operate on a single motion entity and require maintaining self-consistency under textual or temporal instructions, including motion editing~\cite{athanasiou2024motionfix,li2025simmotionedit,jiang2025dynamic,chen2024pay} and temporal completion~\cite{cohan2024flexible,shafir2023human,guo2025motionlab}.
\textit{Heterogeneous} tasks involve multiple motion entities and demand inter-entity coordination, such as reaction generation~\cite{wu2025text2interact,ponce2024in2in,xu2024regennet}. 

Despite the growing diversity of these tasks, existing approaches mainly develop task-specific architectures with dedicated training pipelines, resulting in fragmented solutions that cannot share knowledge across tasks.
A few recent efforts~\cite{guo2025motionlab,luo2024m} attempt to unify multiple tasks within a single framework, yet they are typically limited by insufficient data for unified task learning from scratch and fail to outperform the specialist task models.
In this work, we instead unlock the powerful motion priors in large-scale pretrained T2M models via a lightweight supervised fine-tuning strategy, which effectively adapts a single foundation model to a diverse set of downstream tasks.

\subsection{Text-to-Motion Foundation Models}
Early text-to-motion (T2M) models explored a variety of generative paradigms at relatively small scale, like diffusion-based methods~\cite{tevet2022human,dabral2023mofusion,zhou2024emdm}, masked modeling approaches~\cite{guo2024momask,guo2025snapmogen}, and next-token-prediction~\cite{xiao2025motionstreamer,zhang2023generating}.
However, these models were typically trained on limited datasets (\eg, HumanML3D~\cite{guo2022humanml3d}), restricting the richness of the learned motion priors.
More recently, inspired by scaling laws, current works leverage more scalable architecture designs such as decoder-only transformer~\cite{fan2025go,lu2025scamo,li2026llamo} and diffusion transformer~\cite{lin2025quest,wen2025hy} to achieve impressive generation quality in open-vocabulary settings, substantially scaling up both data and model capacity. 

%% file: section/03_method.tex
\section{Method}
\label{sec:method}
\vspace{-0.3cm}
\subsection{Preliminary}
\label{sec:preliminary}
\vspace{-0.1cm}
\noindent\textbf{Flow Matching.}
We adopt the rectified flow formulation~\cite{lipman2022flow,liu2022flow}.
Let $\mathbf{x}_0 \!\sim\! \mathcal{N}(\mathbf{0}, \mathbf{I})$ be Gaussian noise and $\mathbf{x}_1$ the clean motion.
The interpolation path and target velocity are: $\mathbf{x}_t = (1\!-\!t)\,\mathbf{x}_0 + t\,\mathbf{x}_1,\mathbf{v}_t = \mathbf{x}_1 - \mathbf{x}_0, t \in [0, 1]$.
A neural network $\mathbf{v}_\theta$ is trained to regress this velocity:
$\mathcal{L}_{\text{FM}} = \mathbb{E}_{t,\,\mathbf{x}_0,\,\mathbf{x}_1}\big[\|\mathbf{v}_\theta(\mathbf{x}_t, \mathbf{c}, t) - \mathbf{v}_t\|^2\big]$,
where $\mathbf{c}$ denotes the conditions.
At inference, motion is generated by solving the ODE $\mathrm{d}\mathbf{x}/\mathrm{d}t = \mathbf{v}_\theta(\mathbf{x}_t, \mathbf{c}, t)$ from $t\!=\!0$ to $t\!=\!1$.

\myparagraph{HY-Motion.}
HY-Motion~\cite{wen2025hy} is the first DiT-based motion foundation model pretrained on over 3{,}000 hours of motion data.
It employs a multimodal DiT (MMDiT)~\cite{esser2024scaling} architecture ($460M$/$1B$ parameters) with text conditioning via an LLM encoder (Qwen3-8B~\cite{yang2025qwen3}) and a CLIP encoder~\cite{radford2021learning}.

\myparagraph{Motion Representation.}
Following HY-Motion, each motion is represented as a vector $\mathbf{f}\!\in\!\mathbb{R}^{201}$, comprising global root translation~(3D), root orientation in 6D continuous rotation, 21 local joint rotations~(each 6D), and 22 local joint positions~(each 3D).
All motions are resampled to 30\,fps and normalized to zero mean and unit variance.
A motion sequence of $T$ frames is denoted $\mathbf{x} \in \mathbb{R}^{T \times 201}$.

\subsection{Unified In-Context Motion Generation Formulation}
\label{sec:method_unified_formulation}

Our key insight is: for any target motion sequence to be synthesized, every frame stands in exactly one of three mutually exclusive relationships with respect to the source motion context:
(\romannumeral1)~the frame should be preserved exactly as provided,
(\romannumeral2)~the frame should be generated from scratch, or
(\romannumeral3)~the frame should be edited based on the source motion.
These three relationships are collectively exhaustive: regardless of whether a task involves temporal completion, semantic editing, spatial control, or cross-identity conditioning, the per-frame intent always reduces to one of these three atomic meta-operations.

\myparagraph{Frame-Level Meta-Operation Embeddings.}
We introduce three learnable frame-level meta-operation embeddings, $\texttt{[preserve]}$ ($\texttt{P}$), $\texttt{[generate]}$ ($\texttt{G}$), and $\texttt{[edit]}$ ($\texttt{E}$), each in $\mathbb{R}^{201}$, that represent fine-grained per-frame semantic intent, refining the input embedding of the generative backbone:
$\texttt{[preserve]}$: retain this frame unchanged;
$\texttt{[generate]}$: synthesize this frame from scratch;
$\texttt{[edit]}$: modify the source motion content.
To produce a $T$-frame motion sequence, each frame $i$ is assigned a meta-operation embedding $\tau_i \!\in\! \{\texttt{P}, \texttt{G}, \texttt{E}\}$ and paired with a source motion $\mathbf{s}_i \!\in\! \mathbb{R}^{201}$, set to the original motion $\mathbf{m}_i$ for preservation and editing or to $\mathbf{0}$ when no source reference exists.

\myparagraph{Unified Language Conditioning Design.}
\label{sec:language_instruction}
While the meta-operation embeddings unify what to do at each frame on the motion side, we further unify how to do it on the language conditioning side: all task conditions, from semantic descriptions and editing instructions to precise 3D spatial constraints, are expressed as language and encoded by the same pretrained LLM, without any task-specific conditioning module.
As summarized in Table~\ref{tab:prompt_templates}, we design four prompt templates spanning from natural language to structured specifications:
\textit{Description} serves as the standard semantic interface;
\textit{Editing Instruction} expresses the desired modification;
\textit{Parameterized Trajectory} serializes trajectories into structured language;
and \textit{Spatial Constraint} describes scene-level 3D constraints as compositional language.
This design is also inherently extensible: supporting a new constraint type requires only a new prompt template, with no architectural changes.
\input{table/prompt_templates}

The Parameterized Trajectory template defines a hierarchy of parameterized curves (line, arc, sinusoid, B\'{e}zier, \etc.)\ via a structured format \texttt{\{type:\allowbreak<curve\_type>,\allowbreak\ params:\allowbreak\{...\}\}}, where the parameters uniquely determine the curve's geometric properties such as start/end points, control points, curvature, amplitude, and frequency.
The Spatial Constraint template composes a natural-language sentence specifying the start and goal positions together with a list of obstacles, each described by its center coordinates and safety radius, \eg, ``A person walks from $(x_1, y_1)$ to $(x_2, y_2)$, Avoiding $N$ obstacles at $(x, y, r), \ldots$''.
Both templates encode precise 3D geometric specifications entirely as language, requiring no specialized spatial conditioning module.

\myparagraph{Task Instantiation via Composition.}
Together, as summarized in Table~\ref{tab:task_formulation}, the meta-operation embeddings and language conditions form a complete unified interface: all downstream tasks are fully specified by a per-frame $(\mathbf{s}, \boldsymbol{\tau})$ configuration plus a language condition.
Notably, the three meta-operations form a complete and minimal basis for frame-level intent, and the language conditioning provides an equally open-ended condition space. Any downstream task, including those not yet conceived, can be instantiated by simply defining a new $(\mathbf{s}, \boldsymbol{\tau})$ pattern and a corresponding prompt template, with no architectural changes.
\input{table/task_formulation.tex}

\subsection{In-Context Conditioning Architecture}
\label{sec:architecture_choice}
Given the formulation above, the remaining design question is how to encode and inject the in-context motion signal $(\mathbf{s}, \boldsymbol{\tau})$ into the DiT backbone.
We first describe the encoding process, then compare four injection architectures.

\myparagraph{In-Context Motion Encoder.}
The meta-operation embedding is added to the source to form the task-aware in-context input:
\begin{equation}
  \tilde{\mathbf{s}} = \mathbf{s} + \mathrm{Emb}(\boldsymbol{\tau}),
  \label{eq:frame_token}
\end{equation}
where $\mathbf{s}\!=\![\mathbf{s}_1, \ldots, \mathbf{s}_T]$ and $\boldsymbol{\tau}\!=\![\tau_1, \ldots, \tau_T]$.
The task-aware input $\tilde{\mathbf{s}}$ is then encoded by an MLP encoder $E_{\text{ctx}}$, initialized as a copy of the pretrained input encoder $E_{\text{in}}$ of HY-Motion, inheriting its representation capacity while subsequent finetuning specializes it for noise-free, task-conditioned motion context.

\input{Figure/Architecture_Ablation}
\myparagraph{Architecture Alternatives.}
Given the encoded features $E_{\text{ctx}}(\tilde{\mathbf{s}})$, we consider four representative in-context injection architectures, illustrated in \cref{fig:architecture_ablation}, spanning from input-level fusion to layer-wise modulation:

\noindent(a) \textit{Temporal Fusion} adds $E_{\text{ctx}}(\tilde{\mathbf{s}})$ element-wise to $E_{\text{in}}(\mathbf{x}_t)$ before the DiT blocks, \ie, $\mathbf{x}_t^{\prime} = E_{\text{in}}(\mathbf{x}_t) + E_{\text{ctx}}(\tilde{\mathbf{s}})$, where $\mathbf{x}_t$ denotes the noisy motion at timestep $t$. This fuses in-context motion at the input level with minimal parameter overhead (only 0.207M) while preserving full per-frame granularity.

\noindent(b) \textit{Sequential Concatenation} concatenates $E_{\text{ctx}}(\tilde{\mathbf{s}})$ with $E_{\text{in}}(\mathbf{x}_t)$ along the sequence dimension ($2T$ tokens), with an asymmetric attention mask that makes the in-context motion a read-only conditioning signal. This preserves per-frame granularity but significantly increases computational cost.

\noindent(c) \textit{AdaLN} compresses $E_{\text{ctx}}(\tilde{\mathbf{s}})$ into a global vector via attention pooling and injects it through adaptive layer normalization alongside the timestep embedding. This provides efficient layer-wise modulation but sacrifices per-frame granularity by collapsing the entire source motion into a single vector.

\noindent(d) \textit{ControlNet}~\cite{zhang2023adding} freezes the backbone and clones its double-stream blocks into a parallel trainable branch, injecting in-context motion through zero-initialized residual connections. While this preserves the original backbone weights, it introduces substantial parameter overhead (234M) and restricts the backbone from efficiently adapting to in-context conditioning.

\noindent We systematically ablate all four architectures in Sec.~\ref{sec:architecture_ablation}.

%% file: table/prompt_templates.tex
\begin{table}[t]
    \centering
    \caption{\textbf{Text prompt templates.}
    All task instructions are expressed as text, enabling a unified neural-symbolic interface without any architectural modification.}
    \label{tab:prompt_templates}
    \vspace{-0.2cm}
    \scriptsize
    \setlength{\tabcolsep}{4pt}
    \renewcommand{\arraystretch}{1.15}
    \resizebox{\linewidth}{!}{
    \newcolumntype{R}[1]{>{\raggedright\arraybackslash}p{#1}}
    \begin{tabular}{@{}l@{\hspace{6pt}}R{4.0cm}@{\hspace{6pt}}R{5.8cm}@{}}
    \toprule
    Category & Template & Example \\
    \midrule
    Description
    & \texttt{<motion description>}
    & A man kicks something with his left leg. \\
    \addlinespace[2pt]
    Editing Instruction
    & \texttt{<editing instruction>}
    & {Speed up your motion.} \\
    \addlinespace[2pt]
    Parameterized Trajectory
    & \texttt{\{type:<curve\_type>, params:{attribute1:value1, attribute2:value2, \ldots}\}}
    & {\{type:cubic\_b\'{e}zier, params:\{start:[0.0,0.0], end:[5.22,3.77], P0:[0.0,0.0], P1:[-0.23,3.95], P2:[5.44,-0.17], P3:[5.22,3.77]\}\}} \\
    \addlinespace[2pt]
    Spatial Constraint
    & \texttt{A person walks from ($x_1$,$y_1$) to ($x_2$,$y_2$). Avoiding N obstacles at ($x$,$y$, $r$), \ldots, where $r$ is the safety radius in meters.}
    & {A person walks from (0.00, 0.00) to (3.96, 6.19). Avoiding 3 obstacles at (2.47, 3.04, 0.44), (2.78, 3.82, 0.45), (2.97, 4.68, 0.39), where $r$ is the safety radius in meters.} \\
    \bottomrule
    \end{tabular}}
    \vspace{-0.4cm}
\end{table}

%% file: table/task_formulation.tex
\begin{table}[t]
    \centering
    \caption{\textbf{Unified task formulation.}
    Each task is specified by the per-frame source motion $\mathbf{s}$ and meta-operation embedding $\boldsymbol{\tau}\!\in\!\{\texttt{P},\texttt{G},\texttt{E}\}$.
    $\star$ denotes variant configurations: $(\mathbf{s}\!=\!\mathbf{m}, \texttt{E})$ for editing; $(\mathbf{s}\!=\!\mathbf{0}, \texttt{G})$ for generation.}
    \label{tab:task_formulation}
    \vspace{-0.2cm}
    \scriptsize
    \setlength{\tabcolsep}{2.5pt}
    \renewcommand{\arraystretch}{1.25}
    \newcommand{\cfgl}[1]{\makebox[2.2em][l]{#1}}
    \newcommand{\cfgc}[1]{\makebox[2.2em][c]{#1}}
    \newcommand{\cfgr}[1]{\makebox[2.2em][r]{#1}}
    \resizebox{\linewidth}{!}{
    \begin{tabular}{@{}l@{\hspace{8pt}}c@{\;}c@{\;}c@{\;}c@{\;}c@{\;}c@{\;}c@{\;}c@{\;}c@{}}
    \toprule
    Task & \multicolumn{9}{c}{Configuration ($\mathbf{s}$ top\,/\,$\boldsymbol{\tau}$ bottom)} \\
    \midrule
    \makecell[l]{Text-to-Motion / Trajectory Following /\\ Obstacle Avoidance} & \multicolumn{9}{l}{$s_i \!=\! \mathbf{0},\;\; \tau_i \!=\! \texttt{G} \quad \forall\,i$} \\
    \midrule
    \makecell[l]{Instruction-Based Editing / Reaction /\\ Stylization / Trajectory Following /\\ Obstacle Avoidance} & \multicolumn{9}{l}{$s_i \!=\! \mathbf{m}_i,\;\; \tau_i \!=\! \texttt{E} \quad \forall\,i$} \\
    \midrule
    \multirow{2}{*}{Prediction}
      & \cfgl{$[\mathbf{m}_1,$} & \cfgc{$\ldots,$} & \cfgc{$\mathbf{m}_k,$} & \cfgc{} & \cfgc{} & \cfgc{} & \cfgc{$\star,$} & \cfgc{$\ldots,$} & \cfgr{$\star]$} \\
    & \cfgl{$[\;\texttt{P}\;,$} & \cfgc{$\ldots,$} & \cfgc{$\;\texttt{P}\;,$} & \cfgc{} & \cfgc{} & \cfgc{} & \cfgc{$\star,$} & \cfgc{$\ldots,$} & \cfgr{$\star]$} \\
    \multirow{2}{*}{Backcasting}
      & \cfgl{$[\star,$} & \cfgc{$\ldots,$} & \cfgc{$\star,$} & \cfgc{} & \cfgc{} & \cfgc{} & \cfgc{$\mathbf{m}_{T\text{-}k\text{+}1},$} & \cfgc{$\ldots,$} & \cfgr{$\mathbf{m}_T]$} \\
    & \cfgl{$[\star,$} & \cfgc{$\ldots,$} & \cfgc{$\star,$} & \cfgc{} & \cfgc{} & \cfgc{} & \cfgc{$\;\texttt{P}\;,$} & \cfgc{$\ldots,$} & \cfgr{$\;\texttt{P}\;]$} \\
    \multirow{2}{*}{In-betweening}
      & \cfgl{$[\mathbf{m},$} & \cfgc{$\ldots,$} & \cfgc{$\mathbf{m},$} & \cfgc{$\star,$} & \cfgc{$\ldots,$} & \cfgc{$\star,$} & \cfgc{$\mathbf{m},$} & \cfgc{$\ldots,$} & \cfgr{$\mathbf{m}]$} \\
    & \cfgl{$[\;\texttt{P}\,,$} & \cfgc{$\ldots,$} & \cfgc{$\;\texttt{P}\,,$} & \cfgc{$\star,$} & \cfgc{$\ldots,$} & \cfgc{$\star,$} & \cfgc{$\;\texttt{P}\,,$} & \cfgc{$\ldots,$} & \cfgr{$\;\texttt{P}\;]$} \\
    \multirow{2}{*}{Keyframe Infilling}
      & \cfgl{$[\mathbf{m},$} & \cfgc{$\star,$} & \cfgc{$\ldots,$} & \cfgc{$\star,$} & \cfgc{$\mathbf{m},$} & \cfgc{$\star,$} & \cfgc{$\ldots,$} & \cfgc{$\star,$} & \cfgr{$\mathbf{m}]$} \\
    & \cfgl{$[\;\texttt{P}\,,$} & \cfgc{$\star,$} & \cfgc{$\ldots,$} & \cfgc{$\star,$} & \cfgc{$\;\texttt{P}\,,$} & \cfgc{$\star,$} & \cfgc{$\ldots,$} & \cfgc{$\star,$} & \cfgr{$\;\texttt{P}\;]$} \\
    \bottomrule
    \end{tabular}}
    \vspace{-0.4cm}
    
\end{table}

%% file: Figure/Architecture_Ablation.tex
\begin{figure*}[ht]
    \centering
    \vspace{-0.2cm}
    \includegraphics[width=\textwidth]{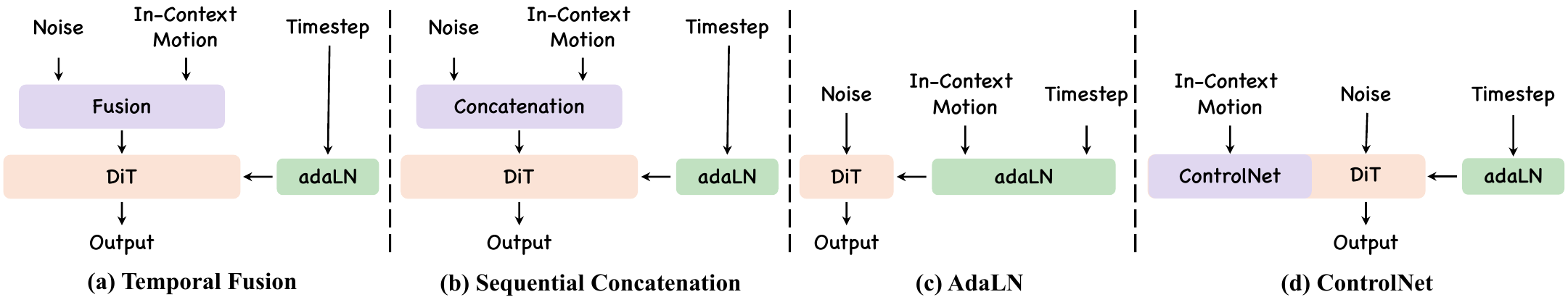}
    \vspace{-0.5cm}
    \caption{Architecture Alternatives for In-Context Motion Conditioning.}
    \label{fig:architecture_ablation}
    \vspace{-0.8cm}
    
\end{figure*}

%% file: section/04_experiment.tex
\section{Experiment}
\label{sec:experiment}

We first ablate the architecture design for injecting in-context motion features (\cref{sec:architecture_ablation}), then systematically evaluate how the generative priors of HY-Motion can be exploited for various downstream tasks and how far these priors can be pushed in more chandelling scenarios, organizing our evaluation along a progressive difficulty axis: in-domain adaptation (\cref{sec:in_domain}), out-of-domain adaptation (\cref{sec:out_of_domain}), and stress tests that probe the boundaries of both the text encoder and the motion backbone (\cref{sec:stress_test}).

\subsection{Experimental Setup}
\label{sec:setup}

\noindent\textbf{Datasets.}
We use HumanML3D~\cite{guo2022humanml3d} for text-to-motion generation and all temporal inpainting subtasks (prediction, backcasting, in-betweening, and keyframe infilling), MotionFix~\cite{athanasiou2024motionfix} for instruction-based motion editing, Inter-X~\cite{xu2024inter} and InterHuman~\cite{liang2024intergen} for reaction generation.
For parameterized trajectory control and obstacle avoidance, we construct a dedicated dataset of 2{,}000 sequences per task.
Details are provided in the supplementary material.

\myparagraph{Evaluation Metrics.}
For text-to-motion generation and temporal inpainting, we follow the standard evaluation protocol~\cite{guo2022humanml3d,guo2024momask,xiao2025motionstreamer} and report: Fr\'{e}chet Inception Distance (FID), measuring the distributional distance between generated and real motions; R-Precision (R@1/2/3) and Multimodal Distance (MM-Dist), measuring text-motion semantic alignment.
For temporal inpainting, we additionally report Mean Per-Joint Position Error (MPJPE, cm) over the full sequence, and $\texttt{[P]}$-MPJPE (cm) restricted to frames assigned $\texttt{[preserve]}$ tokens, measuring how faithfully the model respects the conditioning frames.
For instruction-based motion editing, following~\cite{athanasiou2024motionfix,jiang2025dynamic,li2025simmotionedit}, we report Generated-to-Target retrieval R@1/2/3 and average rank (AvgR) under both batch-level and full-set protocols to evaluate editing accuracy.
For geometric-constrained generation, we report Trajectory Error measuring the average deviation from the target trajectory, inference Latency, and obstacle avoidance Success Rate.

\myparagraph{Implementation Details.}
We employ HY-Motion-Lite (460M)~\cite{wen2025hy} as the pretrained base model and train all models on 4 NVIDIA B200 GPUs with a batch size of 256 and a learning rate of $5 \!\times\! 10^{-5}$.
The unified model is jointly trained on all tasks across all datasets for $100k$ steps, while each expert model is trained on its respective task for $6k$ steps.
During inference, we use a 50-step Euler ODE solver with a classifier-free guidance scale of $2.0$.
All evaluation experiments are conducted on NVIDIA A6000 GPUs.

\myparagraph{Baselines.}
We compare \method with state-of-the-art methods for each task~\cite{tevet2022human,chen2023executing,guo2024momask,zhang2023generating,xiao2025motionstreamer,zhong2023attt2m,jiang2023motiongpt,wen2025hy,jiang2025dynamic,li2025simmotionedit,guo2025motionlab,athanasiou2024motionfix,wan2024tlcontrol,xie2023omnicontrol,cohan2024flexible,pinyoanuntapong2025maskcontrol,karunratanakul2024optimizing,liang2024intergen,ponce2024in2in,yang2025partmotionedit}, all following their official open-source implementations.
Notably, to evaluate the vanilla performance of HY-Motion in temporal inpainting tasks, we implement a baseline based on the classic training-free inversion strategies~\cite{shafir2023human,song2020score}.
Details are provided in the supplementary material.

\subsection{Architecture Ablation}
\label{sec:architecture_ablation}

\input{table/architecture_ablation}
We ablate all four architecture variants described in~\cref{sec:architecture_choice} on the keyframe infilling subtask of temporal inpainting, with quantitative results in \cref{tab:architecture_ablation}.
The results reveal a clear hierarchy.
Temporal Fusion achieves the best $\texttt{[P]}$-MPJPE and FID with minimal overhead ($\Delta$P$=$0.207M, $\Delta$L$=$0.01s), 
balancing faithfulness to the source with responsiveness to the condition. 
Although AdaLN attains the highest R-Precision but critically fails at per-frame constraint, it largely ignores the $\texttt{[preserve]}$ token according to its worst $\texttt{[P]}$-MPJPE among all variants, which validates that compressing the entire source motion into a single global vector fundamentally cannot convey frame-level intent like temporal fusion.
Sequential Concatenation and ControlNet significantly introduce more extra parameters and runtime overhead, yet yield suboptimal performance. 
\input{Figure/keyframe_quality}
The qualitative comparison in \cref{fig:kf_abl} corroborates these findings, where ControlNet exhibits noticeable motion drifting and AdaLN fails to respect the keyframe positions.
To this end, we adopt Temporal Fusion as the default architecture in all subsequent experiments, as it achieves the best trade-off among performance, efficiency, and simplicity.
Hereafter, \method-Expert and \method-Unified denote Temporal Fusion models trained for a single purpose and jointly across all tasks, respectively.
More ablations in various tasks can be found in the supplementary.

%
\input{table/text_to_motion_comparison}
\subsection{In-Domain Task Adaptation}
\label{sec:in_domain}
With Temporal Fusion established as our default architecture, we now evaluate UMO along a progressive difficulty axis, beginning with text-to-motion generation on HumanML3D, which is most closely aligned with the pretraining objective.
As shown in \cref{tab:t2m_hl3d}, directly evaluating HY-Motion on HumanML3D~\cite{guo2022humanml3d} yields high FID despite competitive R-Precision, exposing a distribution gap between its pretraining corpus and HumanML3D rather than a lack of semantic capability, which confirms the necessity of appropriate fine-tuning.
With lightweight adaptation, \method-Expert closes this gap dramatically, demonstrating that fine-tuning successfully activates the semantic priors of the pretrained model.
\method-Unified further advances all metrics, achieving a state-of-the-art FID that surpasses even dedicated text-to-motion models.
This shows a key advantage of our unified multi-task formulation: jointly training across diverse tasks provides complementary supervision that strengthens each individual task, rather than diluting performance through task competition.

\subsection{Out-of-Domain Tasks Adaptation}
\label{sec:out_of_domain}
We evaluate UMO on two categories of out-of-domain tasks, where out-of-domain manifests differently: temporal inpainting challenges the backbone with an in-context generation structure entirely absent from pretraining, while instruction-based editing additionally shifts the language distribution, requiring the text encoder to interpret editing instructions rather than motion descriptions.

\input{table/temporal_inpainting_comparison}
\myparagraph{Temporal Inpainting.}
\label{sec:temporal_inpainting}
\cref{tab:temporal_inpainting} reports results across four subtasks: prediction, backcasting, in-betweening, and keyframe infilling.
Training-free solution~\cite{shafir2023human,song2020score} of HY-Motion fails substantially across all subtasks, confirming that naively repurposing the pretrained model is insufficient.
With proper fine-tuning, both \method-Expert and \method-Unified consistently surpass CondMDI~\cite{cohan2024flexible} and MotionLab~\cite{guo2025motionlab} across all four subtasks in generation quality (FID, R-Precision, MM-Dist).
On frame preservation ($\texttt{[P]}$-MPJPE), \method achieves competitive results through learned $\texttt{[preserve]}$ embeddings alone, without any hard replacement of conditioning frames, which is a fundamentally different mechanism from those used in
CondMDI~\cite{cohan2024flexible} and HY-Motion baseline.
More details are in Appendix.

\input{table/instruction_editing_comparison}
\myparagraph{Instruction-Based Motion Editing.}
\label{sec:instruction_editing}
As shown in \cref{tab:instruction_editing_motionfix}, both \method-Expert and \method-Unified achieve near-perfect batch-level retrieval (R@3=100.0\%), substantially outperforming all prior task-specific methods. 
Under the more challenging full-set protocol, \method-Unified slightly outperforms \method-Expert in R-Precision, 
suggesting that exposure to diverse tasks during joint training enhances the model's fine-grained editing capability.
The qualitative comparison 
\input{Figure/motion_edit_comp}in~\cref{fig:motion_edit} further illustrates that \method applies precise, instruction-aligned edits while faithfully preserving unmodified semantics, confirming the effectiveness of the $\texttt{[edit]}$ embedding in conveying fine-grained editing intent.
Notably, despite never encountering editing instructions during pretraining, the general motion priors can be effectively repurposed for instruction following through our in-context formulation, an early indication of the emergent capability we further explore in the next sections.

\subsection{Stress Test: Ability Emergence}
\label{sec:stress_test}
Beyond out-of-domain adaptation, we further probe the boundaries of pretrained priors with two stress tests that target fundamentally different sources of emergence. 
The first is geometric-constrained generation, where structured spatial specifications are in-domain inputs for pretrained LLMs but represent a completely foreign input distribution to the DiT backbone.
The second is dual-identity reaction generation, a fundamentally different task topology, where single-person motion priors must generalize to two-person interaction. 
In both cases, the question is whether targeted fine-tuning can unlock latent competencies already implicitly embedded in large-scale pretraining.

\input{table/geometric_constraints}
\input{Figure/geo_comp}
\myparagraph{Geometric Constraints.}
\label{sec:geometric_constraints}
As described in \cref{sec:language_instruction}, we encode parameterized trajectories, including line, arc, sinusoid, B\'{e}zier, etc, and spatial constraints, \ie, obstacles, entirely as structured language prompts, encoded by the same pretrained LLM encoder without any specialized spatial conditioning module.
This design introduces a fundamental challenge: while the LLM encoder may have encountered mathematical notation and coordinate data in its pretraining corpus, the DiT backbone has never been exposed to the resulting geometric features as motion conditioning signals.
Consequently, zero-shot evaluation of HY-Motion fails completely (Traj. Err=325.1cm), as the model cannot interpret this entirely new distribution of language features.
We hypothesize that the LLM encoder has already internalized a latent capacity for structured geometric understanding through its pretraining on corpora rich in mathematical notation, code, and coordinate data.
Targeted fine-tuning, in this view, need not teach geometric reasoning from scratch but rather activate this dormant capacity and align it with the DiT's generative space.
\cref{tab:geometric_constraints} strongly confirms our hypothesis. 
Among feed-forward methods, \method-Unified matches OmniControl in trajectory error while being around 90 times faster.
Although optimization-based methods achieve higher precision, they incur latency that is orders of magnitude greater.
\cref{fig:geo_comp} further illustrates that \method produces geometrically faithful motions.
Together, these results suggest that large-scale T2M pretraining implicitly encodes transferable priors that extend into structured spatial reasoning, and since our language conditioning design requires only a new prompt template to incorporate new constraint types, this points to a scalable path toward broader geometric tasks.

\input{table/reaction_generation}
\myparagraph{Dual Identity Reaction Generation.}
\label{sec:dual_identity_reaction_generation}
Our second stress test probes whether single-person motion priors can generalize to two-person interaction, a setting entirely absent from pretraining.
As shown in \cref{tab:reaction_generation}, \method achieves the best FID across all methods, though InterMask~\cite{javed2024intermask} retains a higher R@1, suggesting a trade-off between distributional fidelity and text-motion alignment.
Notably, HY-Motion was never trained on two-person data, yet produces reactions with higher overall motion quality than dedicated multi-person models.
This surprising emergence suggests that core motion priors, such as body dynamics, joint coordination, and temporal coherence, are inherently transferable from single-person to multi-person scenarios, and our in-context formulation can effectively activate them for dual-identity generation.

%% file: table/architecture_ablation.tex
\begin{table}[t]
    \centering
    \caption{Ablation Study on Four Architecture Choices on Keyframe Infilling. $\Delta$P, $\Delta$F, and $\Delta$L denote the additional parameters, FLOPs, and latency over the HY-Motion base model when generating a $360$-frame motion sequence with $50$-step Euler ODE solver. 
    \textbf{Bold} indicates best, \underline{underline} indicates second best. 
    We convert generated results to 272-representation to calculate the FID and R-Precision via official evaluator from MotionStreamer~\cite{xiao2025motionstreamer}.
    We downsample the results from 30 fps to 20 fps to calculate $\texttt{[P]}$-MPJPE to match the evaluation~\cref{tab:temporal_inpainting}.}
    \label{tab:architecture_ablation}
    \vspace{-0.3cm}
    \footnotesize
    \setlength{\tabcolsep}{4pt}
    \renewcommand{\arraystretch}{1.1}
    \resizebox{\columnwidth}{!}{%
    \begin{tabular}{lccccccccc}
    \toprule
    Architectures & $\texttt{[P]}$-MPJPE $\downarrow$ & FID $\downarrow$ & R@1 $\uparrow$ & R@2 $\uparrow$ & R@3 $\uparrow$ & $\Delta$P (M) $\downarrow$ & $\Delta$F (G) $\downarrow$ & $\Delta$L (s) $\downarrow$ \\
    \midrule
    ControlNet & 5.19 & \underline{6.520} & 0.688 & 0.835 & 0.889 & 234.2 & 85.12 & 0.49 \\
    AdaLN & 11.1 & 8.860 & \textbf{0.746} & \textbf{0.875} & \textbf{0.922} & \underline{4.400} & \underline{1.660} & \underline{0.02} \\
    Sequential Concat & \underline{2.04} & 11.77 & 0.680 & 0.831 & 0.887 & \textbf{0.207} & 198.6 & 0.89 \\
    Temporal Fusion & \textbf{0.95} & \textbf{0.476} & \underline{0.692} & \underline{0.842} & \underline{0.896} & \textbf{0.207} & \textbf{0.140} & \textbf{0.01} \\
    \bottomrule
    \end{tabular}%
    }
\end{table}

%% file: Figure/keyframe_quality.tex
\begin{figure*}[ht]
    \centering
    \vspace{-0.1cm}
    
    \includegraphics[width=\textwidth]{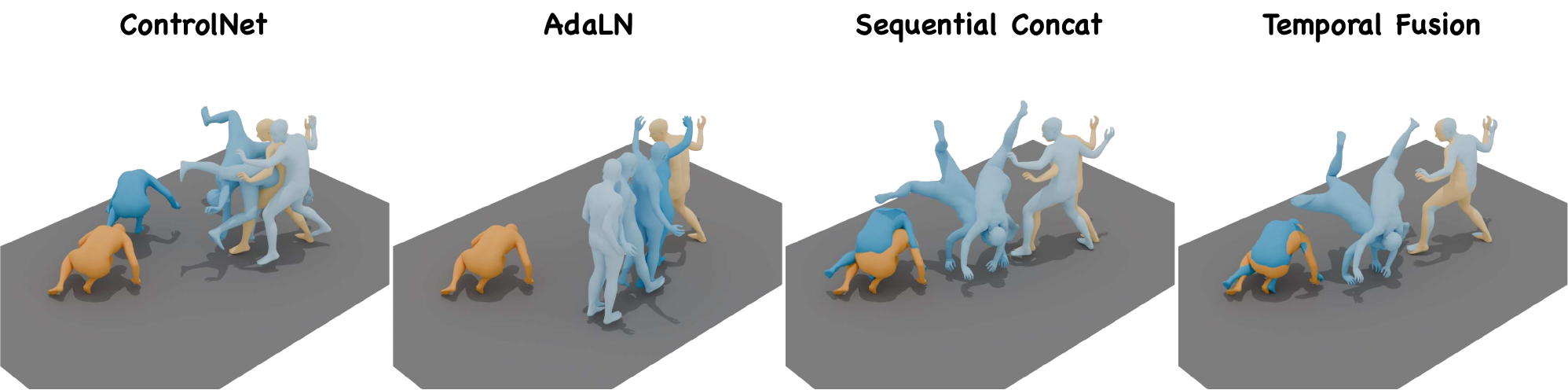}
    \vspace{-0.5cm}
    \caption{Ablation of In-context Conditioning Architectures in Keyframe Infilling.}
    \vspace{-0.3cm}
    
    \label{fig:kf_abl}
\end{figure*}

%% file: table/text_to_motion_comparison.tex
\begin{table}[t]
    \centering
    \caption{Comparison of Text-to-Motion on HumanML3D. \textbf{Bold} indicates best, \underline{underline} indicates second best. We used official evaluator from MotionStreamer~\cite{xiao2025motionstreamer}.}
    \vspace{-0.3cm}
    \label{tab:t2m_hl3d}
    \footnotesize
    \setlength{\tabcolsep}{5pt}
    \renewcommand{\arraystretch}{1.1}
    \resizebox{0.75\textwidth}{!}{
    \begin{tabular}{lccccc}
    \toprule
    Methods & FID $\downarrow$ & R@1 $\uparrow$ & R@2 $\uparrow$ & R@3 $\uparrow$ & MM-D $\downarrow$ \\
    \midrule
    MDM~\cite{tevet2022human} & 23.454 & 0.523 & 0.692 & 0.764 & 17.42 \\
    T2M-GPT~\cite{zhang2023generating} & 12.475 & 0.606 & 0.774 & 0.838 & 16.81 \\
    MoMask~\cite{guo2024momask} & 12.232 & 0.621 & 0.784 & 0.846 & 16.14 \\
    MotionStreamer~\cite{xiao2025motionstreamer} & 11.790 & 0.631 & 0.802 & 0.859 & 16.08 \\
    HY-Motion~\cite{wen2025hy} & 61.035 & 0.667 & 0.818 & 0.876 & 17.53 \\
    \midrule
    \method-Expert & \underline{17.04} & \underline{0.763} & \underline{0.889} & \underline{0.931} & \underline{15.49} \\
    \method-Unified & \textbf{9.460} & \textbf{0.774} & \textbf{0.892} & \textbf{0.933} & \textbf{15.22} \\
    \bottomrule
    \end{tabular}
    }
    \vspace{-0.4cm}
    
\end{table}

%% file: table/temporal_inpainting_comparison.tex
\begin{table}[t]
    \centering
    \caption{Comparison of Temporal Inpainting on HumanML3D based on the MotionLab~\cite{guo2025motionlab} evaluator. We benchmark four subtasks in temporal inpainting.
    \textbf{Bold} indicates best, \underline{underline} indicates second best.
    For fair comparison, we downsample \method and HY-Motion generated results from 30 FPS to 20 FPS and convert them to 263-representation~\cite{guo2022humanml3d} to match the evaluation protocol of all metrics.}
    \vspace{-0.3cm}
    \label{tab:temporal_inpainting}
    \footnotesize
    \setlength{\tabcolsep}{5pt}
    \renewcommand{\arraystretch}{1.1}
    \resizebox{\textwidth}{!}{%
    \begin{tabular}{lccccccc}
    \toprule
    Methods & MPJPE $\downarrow$ & $\texttt{[P]}$-MPJPE $\downarrow$ & FID $\downarrow$ & R@1 $\uparrow$ & R@2 $\uparrow$ & R@3 $\uparrow$ & MM-D $\downarrow$ \\
    \midrule
    \multicolumn{8}{c}{\textcolor{gray}{{\textit{Prediction}}}} \\
    \textcolor{gray}{HY-Motion} & \textcolor{gray}{13.8} & \textcolor{gray}{0.02} & \textcolor{gray}{18.23} & \textcolor{gray}{0.344} & \textcolor{gray}{0.515} & \textcolor{gray}{0.619} & \textcolor{gray}{4.829} \\
    CondMDI~\cite{cohan2024flexible} & 11.2 & \textbf{0.38} & 1.592 & 0.146 & 0.244 & 0.314 & 6.619 \\
    MotionLab~\cite{guo2025motionlab} & 12.3 & 0.81 & 2.844 & 0.179 & 0.283 & 0.374 & 6.170 \\
    \method-Expert & \underline{10.3} & 0.58 & \underline{0.087} & \underline{0.536} & \underline{0.734} & \underline{0.832} & \underline{2.836} \\
    \method-Unified & \textbf{10.2} & \underline{0.54} & \textbf{0.056} & \textbf{0.553} & \textbf{0.744} & \textbf{0.834} & \textbf{2.823} \\
    \midrule
    \multicolumn{8}{c}{\textcolor{gray}{{\textit{Backcasting}}}} \\
    \textcolor{gray}{HY-Motion} & \textcolor{gray}{13.6} & \textcolor{gray}{2.28} & \textcolor{gray}{9.299} & \textcolor{gray}{0.386} & \textcolor{gray}{0.573} & \textcolor{gray}{0.682} & \textcolor{gray}{4.174} \\
    CondMDI~\cite{cohan2024flexible} & 10.9 & 2.18 & 2.408 & 0.131 & 0.226 & 0.297 & 6.895 \\
    MotionLab~\cite{guo2025motionlab} & 12.1 & 3.78 & 1.469 & 0.231 & 0.352 & 0.448 & 5.413 \\
    \method-Expert & \underline{9.80} & \underline{1.66} & \underline{0.182} & \underline{0.521} & \underline{0.728} & \underline{0.814·} & \underline{3.067} \\
    \method-Unified & \textbf{9.55} & \textbf{1.61} & \textbf{0.057} & \textbf{0.563} & \textbf{0.742} & \textbf{0.839} & \textbf{2.808} \\
    \midrule
    \multicolumn{8}{c}{\textcolor{gray}{{\textit{In-Betweening}}}} \\
    \textcolor{gray}{HY-Motion} & \textcolor{gray}{13.7} & \textcolor{gray}{0.97} & \textcolor{gray}{21.97} & \textcolor{gray}{0.319} & \textcolor{gray}{0.474} & \textcolor{gray}{0.577} & \textcolor{gray}{5.282} \\
    CondMDI~\cite{cohan2024flexible} & 10.0 & 1.08 & 0.957 & 0.155 & 0.252 & 0.333 & 6.619 \\
    MotionLab~\cite{guo2025motionlab} & 10.0 & 1.37 & 2.124 & 0.268 & 0.405 & 0.495 & 5.117 \\
    \method-Expert & \underline{8.75} & \underline{0.91} & \underline{0.340} & \underline{0.501} & \underline{0.704} & \underline{0.802} & \underline{3.025} \\
    \method-Unified & \textbf{8.55} & \textbf{0.73} & \textbf{0.050} & \textbf{0.541} & \textbf{0.731} & \textbf{0.827} & \textbf{2.849} \\
    \midrule
    \multicolumn{8}{c}{\textcolor{gray}{{\textit{Keyframe Infilling}}}} \\
    \textcolor{gray}{HY-Motion} & \textcolor{gray}{13.9} & \textcolor{gray}{1.03} & \textcolor{gray}{36.82} & \textcolor{gray}{0.206} & \textcolor{gray}{0.319} & \textcolor{gray}{0.394} & \textcolor{gray}{6.618} \\
    CondMDI~\cite{cohan2024flexible} & 2.72 & \textbf{0.92} & 0.803 & 0.160 & 0.277 & 0.366 & 6.471 \\
    MotionLab~\cite{guo2025motionlab} & 4.01 & 1.44 & 0.117 & 0.495 & 0.673 & 0.776 & 3.131 \\
    \method-Expert & \textbf{2.06} & \underline{0.95} & \underline{0.075} & \underline{0.496} & \underline{0.698} & \underline{0.794} & \underline{3.032} \\
    \method-Unified & \underline{2.67} & \underline{0.95} & \textbf{0.040} & \textbf{0.514} & \textbf{0.707} & \textbf{0.804} & \textbf{2.974} \\
    \bottomrule
    \end{tabular}%
    }
\end{table}

%% file: table/instruction_editing_comparison.tex
\newcommand{\hcell}[1]{\makebox[3.6em][c]{#1}}
\begin{table*}[t]
\centering
\caption{Comparison of Instruction-Based Editing on the MotionFix Dataset. Our method outperforms all other baselines on generated-to-target retrieval. We report R@1, R@2 and R@3 as percentages. $\uparrow$ indicates higher values are better, $\downarrow$ indicates lower values are better, \textbf{bold} indicates best, and \underline{underline} indicates second best.}
\vspace{-0.3cm}
\label{tab:instruction_editing_motionfix}
\small
\setlength{\tabcolsep}{4pt}
\renewcommand{\arraystretch}{1.15}
\resizebox{\textwidth}{!}{
\begin{tabular}{l *{4}{w{c}{3.6em}} @{\hskip 12pt} *{4}{w{c}{3.6em}}}
\toprule
\multirow{2}{*}{Methods} & \multicolumn{4}{c}{\textbf{Generated-to-Target (Batch)}} & \multicolumn{4}{c}{\textbf{Generated-to-Target (Full)}} \\
\cmidrule(lr{10pt}){2-5} \cmidrule(r{10pt}){6-9}
 & \hcell{R@1$\uparrow$} & \hcell{R@2$\uparrow$} & \hcell{R@3$\uparrow$} & \hcell{AvgR$\downarrow$} & \hcell{R@1$\uparrow$} & \hcell{R@2$\uparrow$} & \hcell{R@3$\uparrow$} & \hcell{AvgR$\downarrow$} \\
\midrule
Ground Truth & 100.0 & 100.0 & 100.0 & 1.00 & 64.36 & 88.75 & 95.56 & 1.74 \\
\midrule
\multicolumn{9}{c}{\textcolor{gray}{\textit{Baselines}}} \\
MDM   & 4.03  & 7.56  & 10.48 & 15.55 & 0.10  & 0.10  & 0.10  & --    \\
MDM-BP & 39.10 & 50.09 & 54.84 & 6.46  & 8.69  & 14.71 & 18.36 & 180.99 \\
TMED  & 62.90 & 76.51 & 83.06 & 2.71  & 14.51 & 21.72 & 28.73 & 56.63 \\
SimMotionEdit & 70.62 & 82.92 & 88.12 & 2.38 & 25.49 & 39.33 & 49.21 & 23.49 \\
motionReFit & 66.33 & 80.05 & 84.98 & 2.64 & 14.13 & 23.52 & 30.53 & 54.06 \\
MotionLab & 72.65 & 82.71 & 87.89 & 2.20 & -- & -- & -- & -- \\
PartMotionEdit & 73.96 & 85.83 & 90.21 & 1.92 & 27.27 & 45.06  & 53.36 & 16.24 \\
\midrule
\method-Expert & \textbf{98.08} & \textbf{99.90} & \textbf{100.0} & \textbf{1.02} & \underline{60.91} & \underline{82.13} & \underline{91.31} & \underline{1.76} \\
\method-Unified & \textbf{98.08} & \textbf{99.90} & \textbf{100.0} & \textbf{1.02} & \textbf{61.70} & \textbf{82.23} & \textbf{91.51} & \textbf{1.75} \\
\bottomrule
\end{tabular}}
\vspace{-0.5cm}

\end{table*}

%% file: Figure/motion_edit_comp.tex
\setlength{\intextsep}{0pt}
\begin{wrapfigure}{r}{0.45\textwidth}
    \centering
    \captionsetup{justification=raggedright,singlelinecheck=false}
    \includegraphics[width=0.45\textwidth]{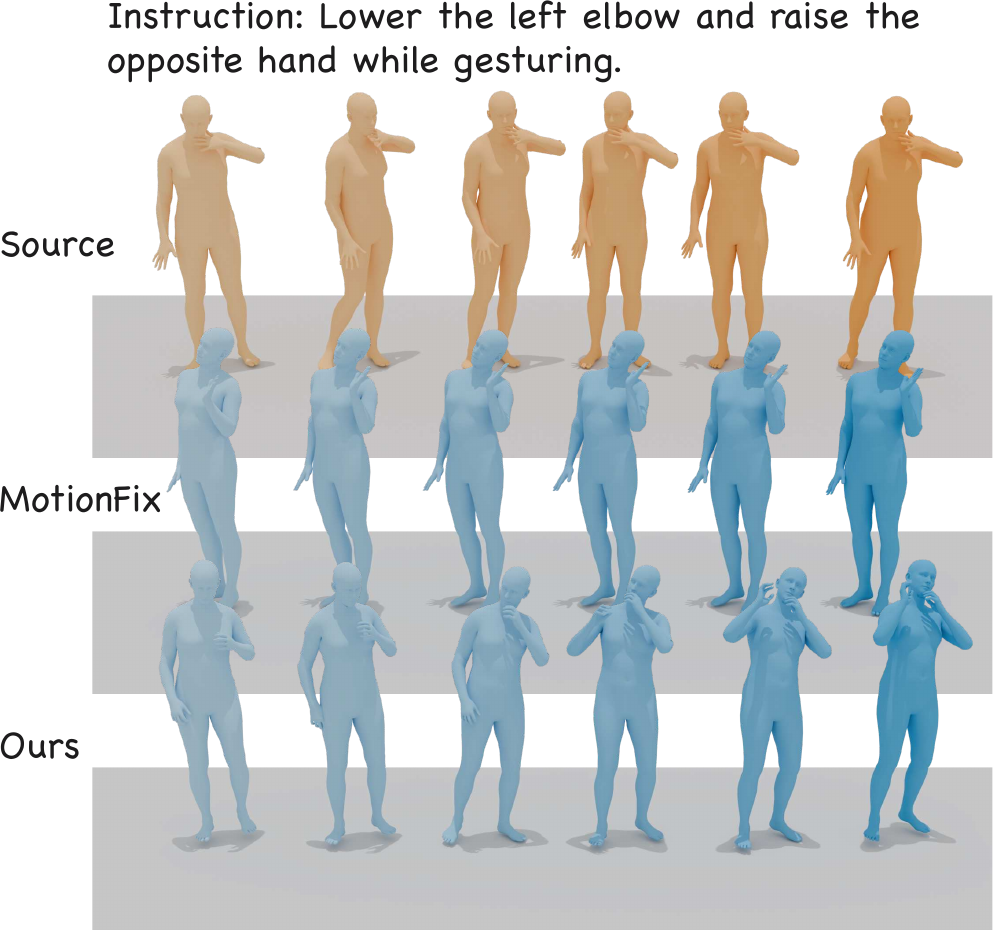}
    \caption{Qualitative Comparison of Text-guided Motion Editing.}
    \vspace{-1cm}
    \label{fig:motion_edit}
\end{wrapfigure}

%% file: table/geometric_constraints.tex
\begin{wraptable}{r}{0.54\textwidth}
    \centering
    \scriptsize
    \setlength{\tabcolsep}{2pt}
    \vspace{-0.4cm}
    
    \renewcommand{\arraystretch}{1.05}
    \captionsetup{justification=raggedright,singlelinecheck=false}
    \caption{Geometric-Constrained Generation. We evaluate trajectory following and obstacle avoidance. Traj.\,Err: average trajectory error (cm). Latency: average inference time (s) per sample. Succ.: obstacle avoidance success rate (\%). \textbf{Bold}: best, \underline{underline}: second best. \colorball{Cerulean}~means feed-forward solutions, \colorball{Orange}~means optimization-based solutions.
    }
    \label{tab:geometric_constraints}
    \begin{tabular}{lccc}
    \toprule
    & Traj.\,Err\,$\downarrow$ & Latency\,$\downarrow$ & Succ.\,$\uparrow$ \\
    \midrule
    \colorball{Cerulean} \textcolor{gray}{HY-Motion} & \textcolor{gray}{325.1} & \textcolor{gray}{0.755} & \textcolor{gray}{--} \\
    \colorball{Cerulean} CondMDI & 70.46 & 38.44 & -- \\
    \colorball{Cerulean} OmniControl & 17.89 & 68.10 & -- \\
    \colorball{Cerulean} TLControl w/o Opt. & 61.15 & \textbf{0.019} & -- \\
    \colorball{Orange} TLControl w/ Opt. & \textbf{2.930} & 3.243 & -- \\
    \colorball{Orange} MaskControl & \underline{3.064} & 31.50 & \underline{0.93} \\
    \colorball{Orange} DNO & -- & 174.9 & 0.88 \\
    \colorball{Cerulean} \method-Expert & 21.29 & \underline{0.759} & 0.92 \\
    \colorball{Cerulean} \method-Unified & 18.78 & \underline{0.759} & \textbf{0.95} \\
    \bottomrule
    \end{tabular}
    \vspace{0.2cm}
    
\end{wraptable}

%% file: Figure/geo_comp.tex
\begin{figure*}[b]
    \centering
    \vspace{-0.5cm}
    
    \includegraphics[width=\textwidth]{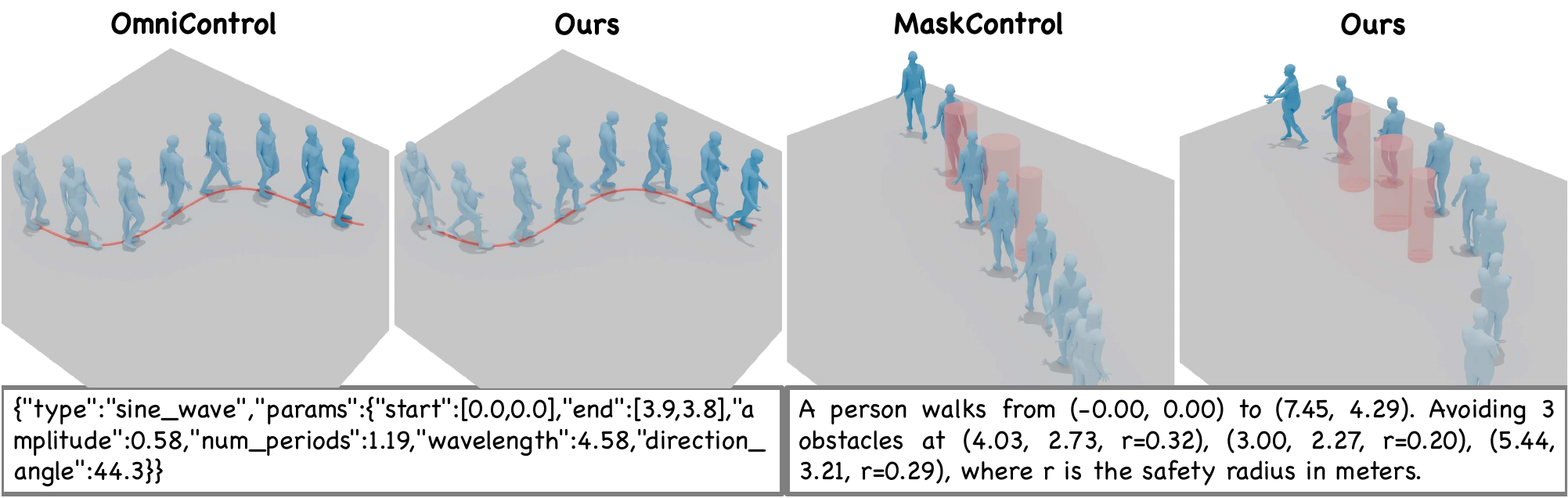}
    \vspace{-0.5cm}
    \caption{Qualitative Comparison of Text-serialized Geometric Constraints.}
    
    \label{fig:geo_comp}
\end{figure*}

%% file: table/reaction_generation.tex
\begin{wraptable}{r}{0.36\textwidth}
    \centering
    \scriptsize
    \setlength{\tabcolsep}{3pt}
    \vspace{-0.4cm}
    \renewcommand{\arraystretch}{1.05}
    \captionsetup{justification=raggedright,singlelinecheck=false}
    \caption{Reaction Generation on InterHuman. \textbf{Bold}: best, \underline{underline}: second best.}
    \label{tab:reaction_generation}
    \begin{tabular}{lcc}
    \toprule
    & FID $\downarrow$ & R@1 $\uparrow$ \\
    \midrule
    InterGen & 52.89 & 0.194 \\
    InterMask & 2.990 & \textbf{0.462} \\
    \method-Expert & \underline{2.130} & 0.428 \\
    \method-Unified & \textbf{2.055} & \underline{0.431} \\
    \bottomrule
    \end{tabular}
    \vspace{0.2cm}
    
\end{wraptable}

%% file: section/05_limitation.tex
\section{Limitations}
\label{sec:limitation}
While \method demonstrates strong generalization across diverse tasks, several limitations remain.
First, the three frame-level meta-operation embeddings operate at the whole-body level and do not support finer-grained part-level control; extending the formulation to a part-specific dimension is a natural direction for future work.
Second, in this paper, we focus on language conditioning and does not handle audio signals such as music or speech. 
Incorporating such modalities requires a multimodal foundation model with audio-language encoding~\cite{xu2025qwen3}.

%% file: section/06_conclusion.tex
\section{Conclusion}
\label{sec:conclusion}
We present \method, a unified in-context adaptation framework that unlocks pretrained motion foundation model priors for diverse downstream tasks within a single model.
The core insight is that any frame-level intent can be expressed as one of three mutually exclusive meta-operations, preserve, generate, or edit, whose compositions compactly specify tasks ranging from temporal inpainting and instruction-based editing to geometric-constrained generation and dual-identity reaction generation.
Equipped with only three learnable embeddings and a lightweight temporal fusion module, \method consistently outperforms task-specific baselines across all benchmarks without any backbone modification.
Two findings are particularly noteworthy: encoding geometric constraints purely in language space achieves results comparable to dedicated spatial conditioning methods, and single-person motion priors transfer effectively to two-person interaction, providing strong evidence that large-scale T2M priors are broadly transferable given an appropriate in-context interface.


%% file: section/x_supp.tex
\appendix


\section{Training-Free Adaptation for HY-Motion.}
\label{sec:finding}

HY-Motion has acquired rich motion priors from large-scale pretraining, yet their potential remains largely confined to a single paradigm: Text-to-Motion (T2M) generation.
We conduct a preliminary experiment to explore whether these priors can be directly repurposed for temporal completion tasks---without any finetuning---by borrowing training-free inversion-based latent manipulation techniques widely used in the vision domain~\cite{mokady2023null,HubermanSpiegelglas2023}.

\paragraph{Inversion-Based Temporal Inpainting.}
We repurpose the pretrained T2M model for temporal inpainting by manipulating its latent denoising trajectory at inference time.
Let $\mathbf{x}_1 = [\mathbf{x}_1^1, \ldots, \mathbf{x}_1^T] \in \mathbb{R}^{T \times 201}$ denote a $T$-frame motion sequence, and let $\mathcal{K} \subset \{1, \ldots, T\}$ with $|\mathcal{K}| = K$ be the index set of known conditioning frames.
Its complement $\mathcal{U} = \{1, \ldots, T\} \setminus \mathcal{K}$ denotes the frames to be synthesized.
Starting from $t\!=\!0$, all $T$ frames are initialized as $\boldsymbol{\epsilon}^i \!\sim\! \mathcal{N}(\mathbf{0}, \mathbf{I})$.
At each denoising step $t$, we replace the tokens at known positions $\mathcal{K}$ with their noise-level--matched counterparts derived from the clean frames:
\begin{equation}
  \hat{\mathbf{x}}_t^i \leftarrow (1 - t)\,\boldsymbol{\epsilon}^i + t\,\mathbf{x}_1^i, \quad i \in \mathcal{K},
  \label{eq:inversion}
\end{equation}
where $\boldsymbol{\epsilon}^i$ is the same noise sample drawn at $t\!=\!0$ for frame $i$.
This effectively anchors each known frame, ensuring its noise level is consistent with the surrounding tokens being denoised.
The DiT then jointly denoises all $T$ tokens to synthesize temporally coherent completions for $\mathcal{U}$ while respecting the anchored conditioning frames $\mathcal{K}$.

As shown in \cref{tab:temporal_inpainting_supp}, this training-free approach, which is successful in the vision domain, does not transfer gracefully to the motion domain.
It yields incoherent and degraded motions far from satisfactory, indicating that the anchored frames provide very limited signals to effectively guide the generation.
To truly unlock the motion priors inside the pretrained foundation model for diverse downstream tasks, training-free latent manipulation is insufficient---proper finetuning is necessary.

\section{Full Quantitative Results}
\label{sec:full_results}

We provide the complete quantitative results with all four architecture variants (Temporal Fusion, Sequential Concatenation, AdaLN, ControlNet) alongside baselines for text-to-motion generation (\cref{tab:t2m_hl3d_supp}), temporal inpainting (\cref{tab:temporal_inpainting_supp}), and instruction-based motion editing (\cref{tab:instruction_editing_motionfix_supp}).

These full results complement the keyframe infilling ablation in the main paper and consistently corroborate the same conclusion across all tasks:
{Temporal Fusion achieves the best overall trade-off.}
Across all temporal inpainting subtasks (\cref{tab:temporal_inpainting_supp}), Temporal Fusion achieves the lowest MPJPE and $\texttt{[P]}$-MPJPE by a large margin, confirming its superior ability to respect per-frame conditioning constraints.
AdaLN occasionally attains higher R-Precision but consistently exhibits the worst $\texttt{[P]}$-MPJPE, validating that compressing the source motion into a single global vector fundamentally sacrifices frame-level fidelity.
ControlNet and Sequential Concatenation introduce substantially more parameters and latency yet yield no quality advantage.

\input{table/text_to_motion_comparison_supp}

The results further demonstrate that {proper fine-tuning successfully unlocks the generative priors of HY-Motion.}
Across all tasks, training-free or zero-shot application of HY-Motion fails substantially, whereas \method variants dramatically surpass all prior task-specific state-of-the-art methods.
Notably, \method-Unified consistently matches or outperforms \method-Expert on every benchmark, demonstrating that multi-task joint training provides complementary supervision that strengthens individual tasks.

\input{table/instruction_editing_comparison_supp}
\input{table/meta_operation_ablation}

\section{Ablation on the Necessity of The Three Meta-Operation Embeddings}
\label{sec:meta_op_ablation}

We validate that all three meta-operation embeddings, $\texttt{[preserve]}$, $\texttt{[generate]}$, and $\texttt{[edit]}$, are necessary.
Specifically, we collapse the token set from $\{\texttt{P}, \texttt{G}, \texttt{E}\}$ to $\{\texttt{P}, \texttt{G}\}$ by removing $\texttt{[edit]}$ and merging its role into $\texttt{[generate]}$.
Frames that previously received $\texttt{[edit]}$ are instead assigned $\texttt{[generate]}$ with the source motion still provided as input.
All variants use the same Temporal Fusion architecture and identical hyperparameters.

As shown in \cref{tab:meta_op_ablation}, removing $\texttt{[edit]}$ degrades performance on both temporal inpainting and instruction-based editing.
The key reason is that $\texttt{[generate]}$ must now additionally absorb the editing semantics previously carried by $\texttt{[edit]}$, forcing a single token to handle both unconditional synthesis and source-aware modification.
This overloaded responsibility dilutes the precision of $\texttt{[generate]}$, degrading temporal inpainting.
For instruction-based editing, the impact is even more direct: without $\texttt{[edit]}$, the model has no explicit editing signal, and instead over-generates from scratch, disregarding the source motion.

\section{Geometric-Constrained Dataset Construction.}
\label{sec:dataset_details}
For the parameterized trajectory following and obstacle avoidance task, no suitable public dataset exists.
We construct a dedicated dataset of 2{,}000 sequences per task using a three-stage pipeline.

\subsection{Parameterized Trajectory Generation.}
We define a hierarchy of parameterized curves organized into three complexity levels:

\noindent\textit{Level~1 -- Linear:} A straight-line trajectory $P(t) = P_0 + t(P_1 - P_0)$, parameterized by start point $P_0$, end point $P_1$, and walking speed.

\noindent\textit{Level~2 -- Arc:} A circular arc $P(\theta) = c + r[\cos\theta, 0, \sin\theta]$, parameterized by center $c$, radius $r$ ($\geq 2.0$\,m), turn angle ($\frac{\pi}{2}$ to $\pi$), and direction (clockwise/counter-clockwise).

\noindent\textit{Level~3 -- Inflection:} Let the {chord} denote the straight line segment connecting the start point $P_0$ and the end point $P_{\text{end}}$, with chord length $\ell = \|P_{\text{end}} - P_0\|$.
(a) Quadratic B\'{e}zier $B(t) = (1\!-\!t)^2 P_0 + 2(1\!-\!t)t\, P_1 + t^2 P_2$, with lateral control-point offset $0.3$--$0.6{\times}\ell$;
(b) Cubic B\'{e}zier with S-shaped control points;
(c) Sinusoidal perturbation $A\sin(2\pi f t)$ applied perpendicular to the chord.
The three curve types are sampled with equal probability.

All trajectories are constrained to a maximum of 190 frames (6.33\,s at 30\,fps) and an arc length of 1.0--8.0\,m.
Each trajectory is canonicalized by translating the start to the origin and rotating around the $Y$-axis.
For each curve, we record the full set of defining parameters (\eg, start/end points, radius, control points, amplitude, frequency).

\subsection{Motion Synthesis via TLControl.}
Each parameterized trajectory is used as the pelvis trajectory to drive TLControl~\cite{wan2024tlcontrol}, a trajectory-conditioned motion generation model, which synthesizes a physically plausible full-body motion sequence that follows the prescribed root path.
The generated skeleton sequences are then converted to SMPL+H parameters via inverse kinematics.

\subsection{Training Data Assembly.}
From the synthesized motions, we assemble training pairs for each task:

\noindent\textit{Trajectory Following:}
We randomly generate 200 Level~1, 800 Level~2, and 1{,}000 Level~3 trajectories and record the corresponding curve parameters.
The source motion is a straight-line walk between the start and end points of the parameterized curve.
The target motion is the synthesized motion following the parameterized trajectory.
The text condition is a structured parameterized trajectory template (cf.\ Table~1 of the main paper).
We prepare two versions of templates for each curve type: a \emph{full} version that records every defining parameter, and a \emph{minimal} version that retains only the smallest set of parameters sufficient to uniquely determine the curve.
We adopt the minimal version for training for two reasons: (1)~the text encoder of HY-Motion has a limited maximum token length, and the full template may exceed this budget; (2)~the minimal parameterization serves as a stricter evaluation of whether the model truly understands each geometric constraint, since every parameter in the prompt is essential and must be faithfully followed.
The two versions are listed below:

\noindent\textbf{Full templates:}
\begin{itemize}
  \item \{type: linear, params: \{start: [..], end: [..], speed: [..], chord\_len: [..]\}\}
  \item \{type: arc, params: \{start: [..], end: [..], center: [..], radius: [..], angle: [..], dir: [..], arc\_len: [..]\}\}
  \item \{type: quad\_b\'{e}zier, params: \{start: [..], end: [..], P1: [..], chord\_len: [..], offset\_ratio: [..]\}\}
  \item \{type: cubic\_b\'{e}zier, params: \{start: [..], end: [..], P1: [..], P2: [..], chord\_len: [..]\}\}
  \item \{type: sinusoidal, params: \{start: [..], end: [..], A: [..], f: [..], chord\_len: [..]\}\}
\end{itemize}

\noindent\textbf{Minimal templates (used for training):}
\begin{itemize}
  \item \{type: linear, params: \{start: [..], end: [..], speed: [..]\}\}
  \item \{type: arc, params: \{start: [..], end: [..], center: [..], dir: [..]\}\}
  \item \{type: quad\_b\'{e}zier, params: \{start: [..], end: [..], P1: [..]\}\}
  \item \{type: cubic\_b\'{e}zier, params: \{start: [..], end: [..], P1: [..], P2: [..]\}\}
  \item \{type: sinusoidal, params: \{start: [..], end: [..], A: [..], f: [..]\}\}
\end{itemize}

\noindent\textit{Obstacle Avoidance:}
We randomly generate 100 Level~1, 900 Level~2, and 1{,}000 Level~3 trajectories.
Building on the trajectory-following data, we construct obstacle avoidance scenarios via curvature-driven obstacle placement.
For each parameterized curve, we compute its curvature profile and detect turning points via curvature peak detection.
Obstacles are placed in the high-curvature regions on the \emph{inner side} of turns, ensuring that the obstacles causally explain why the trajectory curves.
The number of obstacles scales with trajectory complexity: 1--3 for Levels~1--2, and 3--5 for Level~3.
The text condition uses the spatial constraint template (cf.\ Table~1 of the main paper), specifying start/goal positions and a list of obstacles with center coordinates and safety radii.

\section{Qualitative Results}
\label{sec:more_qualitative}

As shown in \cref{fig:qual_prediction,fig:qual_backcasting,fig:qual_inbetween,fig:qual_infilling,fig:qual_motionediting,fig:qual_t2m,fig:qual_traj,fig:qual_obstacle,fig:qual_react}, we present additional qualitative results of \method on diverse tasks.

\input{table/temporal_inpainting_comparison_supp}

\clearpage

\input{Figure/qual_t2m}
\input{Figure/qual_prediction}
\input{Figure/qual_backcasting}
\input{Figure/qual_inbetween}
\input{Figure/qual_infilling}
\input{Figure/qual_motionediting}
\input{Figure/qual_traj}
\input{Figure/qual_obstacle}
\input{Figure/qual_react}

%% file: table/text_to_motion_comparison_supp.tex
\begin{table}[t]
    \centering
    \caption{{Comparison of Text-to-Motion on HumanML3D.} \textbf{Bold} indicates best, \underline{underline} indicates second best. We used official evaluator from MotionStreamer~\cite{xiao2025motionstreamer}.}
    \label{tab:t2m_hl3d_supp}
    \footnotesize
    \setlength{\tabcolsep}{5pt}
    \renewcommand{\arraystretch}{1.1}
    \begin{tabular}{lccccc}
    \toprule
    Methods & FID $\downarrow$ & R@1 $\uparrow$ & R@2 $\uparrow$ & R@3 $\uparrow$ & MM-D $\downarrow$ \\
    \midrule
    \multicolumn{6}{c}{\textcolor{gray}{\textit{Baselines}}} \\
    MDM~\cite{tevet2022human} & 23.454 & 0.523 & 0.692 & 0.764 & 17.42 \\
    MLD~\cite{chen2023executing} & 18.236 & 0.546 & 0.730 & 0.792 & 16.64 \\
    T2M-GPT~\cite{zhang2023generating} & 12.475 & 0.606 & 0.774 & 0.838 & 16.81 \\
    MotionGPT~\cite{jiang2023motiongpt} & 14.375 & 0.456 & 0.598 & 0.628 & 17.89 \\
    AttT2M~\cite{zhong2023attt2m} & 15.428 & 0.592 & 0.765 & 0.834 & 15.73 \\
    MoMask~\cite{guo2024momask} & 12.232 & 0.621 & 0.784 & 0.846 & 16.14 \\
    MotionStreamer~\cite{xiao2025motionstreamer} & 11.790 & 0.631 & 0.802 & 0.859 & 16.08 \\
    HY-Motion~\cite{wen2025hy} & 61.035 & 0.667 & 0.818 & 0.876 & 17.53 \\
    \midrule
    \textcolor{gray}{Architectures} & \multicolumn{5}{c}{\kern-1.9cm\textcolor{gray}{\textit{\method-Expert}}\kern1.5cm} \\
    ControlNet & 26.22 & 0.703 & 0.841 & 0.895 & 16.44 \\
    AdaLN & 18.12 & 0.754 & 0.883 & 0.931 & 15.55 \\
    Sequential Concat & 19.40 & 0.755 & 0.876 & 0.923 & 15.68 \\
    Temporal Fusion & 17.04 & 0.763 & 0.889 & 0.931 & 15.49 \\
    \cdashline{1-6}
    \multicolumn{6}{c}{\textcolor{gray}{\textit{\method-Unified}}} \\
    Temporal Fusion & 9.460 & 0.774 & 0.892 & 0.933 & 15.22 \\
    \bottomrule
    \end{tabular}
\end{table}

%% file: table/instruction_editing_comparison_supp.tex
\begin{table*}[t]
\centering
\caption{Comparison of Instruction-Based Editing on the MotionFix Dataset. Our method outperforms all other baselines on generated-to-target retrieval. We report R@1, R@2 and R@3 as percentages. $\uparrow$ indicates higher values are better, $\downarrow$ indicates lower values are better, \textbf{bold} indicates best, and \underline{underline} indicates second best.}
\label{tab:instruction_editing_motionfix_supp}
\small
\setlength{\tabcolsep}{4pt}
\renewcommand{\arraystretch}{1.15}
\resizebox{\textwidth}{!}{
\begin{tabular}{l *{4}{w{c}{3.6em}} @{\hskip 12pt} *{4}{w{c}{3.6em}}}
\toprule
\multirow{2}{*}{Methods} & \multicolumn{4}{c}{\textbf{Generated-to-Target (Batch)}} & \multicolumn{4}{c}{\textbf{Generated-to-Target (Full)}} \\
\cmidrule(lr{10pt}){2-5} \cmidrule(r{10pt}){6-9}
 & \hcell{R@1$\uparrow$} & \hcell{R@2$\uparrow$} & \hcell{R@3$\uparrow$} & \hcell{AvgR$\downarrow$} & \hcell{R@1$\uparrow$} & \hcell{R@2$\uparrow$} & \hcell{R@3$\uparrow$} & \hcell{AvgR$\downarrow$} \\
\midrule
Ground Truth & 100.0 & 100.0 & 100.0 & 1.00 & 64.36 & 88.75 & 95.56 & 1.74 \\
\midrule
\multicolumn{9}{c}{\textcolor{gray}{\textit{Baselines}}} \\
MDM   & 4.03  & 7.56  & 10.48 & 15.55 & 0.10  & 0.10  & 0.10  & --    \\
MDM-BP & 39.10 & 50.09 & 54.84 & 6.46  & 8.69  & 14.71 & 18.36 & 180.99 \\
TMED  & 62.90 & 76.51 & 83.06 & 2.71  & 14.51 & 21.72 & 28.73 & 56.63 \\
SimMotionEdit & 70.62 & 82.92 & 88.12 & 2.38 & 25.49 & 39.33 & 49.21 & 23.49 \\
motionReFit & 66.33 & 80.05 & 84.98 & 2.64 & 14.13 & 23.52 & 30.53 & 54.06 \\
MotionLab & 72.65 & 82.71 & 87.89 & 2.20 & -- & -- & -- & -- \\
PartMotionEdit & 73.96 & 85.83 & 90.21 & 1.92 & 27.27 & 45.06  & 53.36 & 16.24 \\
\midrule
\textcolor{gray}{Architectures} & \multicolumn{8}{c}{\kern-1.5cm\textcolor{gray}{\textit{\method-Expert}}\kern1.5cm} \\
ControlNet & 95.77 & \underline{99.29} & \underline{99.70} & \underline{1.06} & 51.63 & 72.16 & 81.84 & 2.97 \\
AdaLN & \underline{97.88} & \textbf{99.90} & \textbf{100.0} & \textbf{1.02} & 60.61 & 81.15 & 90.42 & 1.82 \\
Sequential Concat & 93.65 & 98.19 & 99.09 & 1.17 & 44.32 & 63.08 & 73.25 & 6.53 \\
Temporal Fusion & \textbf{98.08} & \textbf{99.90} & \textbf{100.0} & \textbf{1.02} & \underline{60.91} & \underline{82.13} & \underline{91.31} & \underline{1.76} \\
\midrule
\multicolumn{9}{c}{\textcolor{gray}{\textit{\method-Unified}}} \\
Temporal Fusion & \textbf{98.08} & \textbf{99.90} & \textbf{100.0} & \textbf{1.02} & \textbf{61.70} & \textbf{82.23} & \textbf{91.51} & \textbf{1.75} \\
\bottomrule
\end{tabular}}
\end{table*}

%% file: table/meta_operation_ablation.tex
\begin{table}[t]
    \centering
    \caption{Ablation on the meta-operation embedding set.
    We compare the reduced two-token set $\{\texttt{P}, \texttt{G}\}$ (merging $\texttt{[edit]}$ into $\texttt{[generate]}$) against the full three-token set $\{\texttt{P}, \texttt{G}, \texttt{E}\}$ on temporal inpainting (top) and instruction-based editing (bottom).
    Removing $\texttt{[edit]}$ consistently degrades performance across all tasks, confirming the necessity of all three meta-operations.
    \textbf{Bold} indicates best.}
    \label{tab:meta_op_ablation}
    \footnotesize
    \setlength{\tabcolsep}{5pt}
    \renewcommand{\arraystretch}{1.1}
    \resizebox{\textwidth}{!}{%
    \begin{tabular}{llccccccc}
    \toprule
    Task & Token Set & MPJPE $\downarrow$ & $\texttt{[P]}$-MPJPE $\downarrow$ & FID $\downarrow$ & R@1 $\uparrow$ & R@2 $\uparrow$ & R@3 $\uparrow$ & MM-D $\downarrow$ \\
    \midrule
    \multirow{2}{*}{Prediction}
    & $\{\texttt{P}, \texttt{G}\}$ & \textbf{10.2} & 0.71 & 0.058 & 0.550 & 0.742 & 0.831 & 2.849 \\
    & $\{\texttt{P}, \texttt{G}, \texttt{E}\}$ & \textbf{10.2} & \textbf{0.54} & \textbf{0.056} & \textbf{0.553} & \textbf{0.744} & \textbf{0.834} & \textbf{2.823} \\
    \midrule
    \multirow{2}{*}{Backcasting}
    & $\{\texttt{P}, \texttt{G}\}$ & \textbf{9.53} & 1.77 & \textbf{0.056} & 0.550 & 0.740 & 0.836 & 2.816 \\
    & $\{\texttt{P}, \texttt{G}, \texttt{E}\}$ & 9.55 & \textbf{1.61} & 0.057 & \textbf{0.563} & \textbf{0.742} & \textbf{0.839} & \textbf{2.808} \\
    \midrule
    \multirow{2}{*}{In-Between.}
    & $\{\texttt{P}, \texttt{G}\}$ & \textbf{8.52} & 0.92 & \textbf{0.049} & 0.536 & 0.727 & 0.808 & 2.851 \\
    & $\{\texttt{P}, \texttt{G}, \texttt{E}\}$ & 8.55 & \textbf{0.73} & 0.050 & \textbf{0.541} & \textbf{0.731} & \textbf{0.827} & \textbf{2.849} \\
    \midrule
    \multirow{2}{*}{Infilling}
    & $\{\texttt{P}, \texttt{G}\}$ & 2.74 & 1.18 & 0.050 & 0.503 & 0.699 & 0.797 & 3.004 \\
    & $\{\texttt{P}, \texttt{G}, \texttt{E}\}$ & \textbf{2.67} & \textbf{0.95} & \textbf{0.040} & \textbf{0.514} & \textbf{0.707} & \textbf{0.804} & \textbf{2.974} \\
    \bottomrule
    \end{tabular}%
    }

    \vspace{0.25cm}

    \resizebox{\textwidth}{!}{%
    \begin{tabular}{ll*{4}{c}@{\hskip 8pt}*{4}{c}}
    \toprule
    \multirow{2}{*}{Task} & \multirow{2}{*}{Token Set} & \multicolumn{4}{c}{Gen-to-Target (Batch)} & \multicolumn{4}{c}{Gen-to-Target (Full)} \\
    \cmidrule(lr){3-6} \cmidrule(lr){7-10}
     & & R@1 $\uparrow$ & R@2 $\uparrow$ & R@3 $\uparrow$ & AvgR $\downarrow$ & R@1 $\uparrow$ & R@2 $\uparrow$ & R@3 $\uparrow$ & AvgR $\downarrow$ \\
    \midrule
    \multirow{2}{*}{Editing}
    & $\{\texttt{P}, \texttt{G}\}$ & 97.98 & 99.80 & 99.90 & 1.03 & 61.60 & 81.33 & 91.41 & 1.97 \\
    & $\{\texttt{P}, \texttt{G}, \texttt{E}\}$ & \textbf{98.08} & \textbf{99.90} & \textbf{100.0} & \textbf{1.02} & \textbf{61.70} & \textbf{82.23} & \textbf{91.51} & \textbf{1.75} \\
    \bottomrule
    \end{tabular}%
    }
\end{table}

%% file: table/temporal_inpainting_comparison_supp.tex
\begin{table}[t]
    \centering
    \caption{Comparison of Temporal Inpainting on HumanML3D with all architecture variants. We benchmark four subtasks using the MotionLab~\cite{guo2025motionlab} evaluator.
    \textbf{Bold} indicates best, \underline{underline} indicates second best.
    For fair comparison, we downsample \method and HY-Motion generated results from 30 FPS to 20 FPS and convert them to 263-representation~\cite{guo2022humanml3d} to match the evaluation protocol of all metrics.}
    \label{tab:temporal_inpainting_supp}
    \footnotesize
    \setlength{\tabcolsep}{5pt}
    \renewcommand{\arraystretch}{1.1}
    \resizebox{0.95\textwidth}{!}{%
    \begin{tabular}{lccccccc}
    \toprule
    Methods & MPJPE $\downarrow$ & $\texttt{[P]}$-MPJPE $\downarrow$ & FID $\downarrow$ & R@1 $\uparrow$ & R@2 $\uparrow$ & R@3 $\uparrow$ & MM-D $\downarrow$ \\
    \midrule
    \multicolumn{8}{c}{\textcolor{gray}{{\textit{Prediction}}}} \\
    \textcolor{gray}{HY-Motion} & \textcolor{gray}{13.8} & \textcolor{gray}{0.02} & \textcolor{gray}{18.23} & \textcolor{gray}{0.344} & \textcolor{gray}{0.515} & \textcolor{gray}{0.619} & \textcolor{gray}{4.829} \\
    CondMDI~\cite{cohan2024flexible} & 11.2 & \textbf{0.38} & 1.592 & 0.146 & 0.244 & 0.314 & 6.619 \\
    MotionLab~\cite{guo2025motionlab} & 12.3 & 0.81 & 2.844 & 0.179 & 0.283 & 0.374 & 6.170 \\
    \cdashline{1-8}
    \textcolor{gray}{Architectures} & \multicolumn{7}{c}{\kern-1.5cm\textcolor{gray}{\textit{\method-Expert}}\kern1.5cm} \\
    ControlNet & 11.3 & 2.99 & 0.526 & 0.491 & 0.690 & 0.802 & 3.055 \\
    AdaLN & 10.8 & 7.75 & 0.128 & 0.535 & 0.733 & \underline{0.833} & \underline{2.830} \\
    Sequential Concat & 11.2 & 0.79 & 1.437 & 0.496 & 0.683 & 0.782 & 3.115 \\
    Temporal Fusion & \underline{10.3} & 0.58 & \underline{0.087} & \underline{0.536} & \underline{0.734} & 0.832 & 2.836 \\
    \cdashline{1-8}
    \multicolumn{8}{c}{\textcolor{gray}{\textit{\method-Unified}}} \\
    Temporal Fusion & \textbf{10.2} & \underline{0.54} & \textbf{0.056} & \textbf{0.553} & \textbf{0.744} & \textbf{0.834} & \textbf{2.823} \\
    \midrule
    \multicolumn{8}{c}{\textcolor{gray}{{\textit{Backcasting}}}} \\
    \textcolor{gray}{HY-Motion} & \textcolor{gray}{13.6} & \textcolor{gray}{2.28} & \textcolor{gray}{9.299} & \textcolor{gray}{0.386} & \textcolor{gray}{0.573} & \textcolor{gray}{0.682} & \textcolor{gray}{4.174} \\
    CondMDI~\cite{cohan2024flexible} & 10.9 & 2.18 & 2.408 & 0.131 & 0.226 & 0.297 & 6.895 \\
    MotionLab~\cite{guo2025motionlab} & 12.1 & 3.78 & 1.469 & 0.231 & 0.352 & 0.448 & 5.413 \\
    \cdashline{1-8}
    \textcolor{gray}{Architectures} & \multicolumn{7}{c}{\kern-1.5cm\textcolor{gray}{\textit{\method-Expert}}\kern1.5cm} \\
    ControlNet & 11.4 & 2.73 & 0.605 & 0.478 & 0.686 & 0.795 & 3.109 \\
    AdaLN & 11.1 & 6.23 & \underline{0.136} & \underline{0.536} & \underline{0.739} & \underline{0.833} & \underline{2.842} \\
    Sequential Concat & 10.3 & 1.89 & 2.153 & 0.444 & 0.640 & 0.749 & 3.418 \\
    Temporal Fusion & \underline{9.80} & \underline{1.66} & 0.182 & 0.521 & 0.728 & 0.814 & 3.067 \\
    \cdashline{1-8}
    \multicolumn{8}{c}{\textcolor{gray}{\textit{\method-Unified}}} \\
    Temporal Fusion & \textbf{9.55} & \textbf{1.61} & \textbf{0.057} & \textbf{0.563} & \textbf{0.742} & \textbf{0.839} & \textbf{2.808} \\
    \midrule
    \multicolumn{8}{c}{\textcolor{gray}{{\textit{In-Betweening}}}} \\
    \textcolor{gray}{HY-Motion} & \textcolor{gray}{13.7} & \textcolor{gray}{0.97} & \textcolor{gray}{21.97} & \textcolor{gray}{0.319} & \textcolor{gray}{0.474} & \textcolor{gray}{0.577} & \textcolor{gray}{5.282} \\
    CondMDI~\cite{cohan2024flexible} & 10.0 & 1.08 & 0.957 & 0.155 & 0.252 & 0.333 & 6.619 \\
    MotionLab~\cite{guo2025motionlab} & 10.0 & 1.37 & 2.124 & 0.268 & 0.405 & 0.495 & 5.117 \\
    \cdashline{1-8}
    \textcolor{gray}{Architectures} & \multicolumn{7}{c}{\kern-1.5cm\textcolor{gray}{\textit{\method-Expert}}\kern1.5cm} \\
    ControlNet & 11.3 & 2.21 & 0.558 & 0.486 & 0.692 & 0.793 & 3.088 \\
    AdaLN & 10.6 & 5.05 & \underline{0.112} & \underline{0.530} & \textbf{0.735} & \textbf{0.836} & \textbf{2.834} \\
    Sequential Concat & 9.82 & 1.50 & 0.583 & 0.488 & 0.681 & 0.788 & 3.079 \\
    Temporal Fusion & \underline{8.75} & \underline{0.91} & 0.340 & 0.501 & 0.704 & 0.802 & 3.025 \\
    \cdashline{1-8}
    \multicolumn{8}{c}{\textcolor{gray}{\textit{\method-Unified}}} \\
    Temporal Fusion & \textbf{8.55} & \textbf{0.73} & \textbf{0.050} & \textbf{0.541} & \underline{0.731} & \underline{0.827} & \underline{2.849} \\
    \midrule
    \multicolumn{8}{c}{\textcolor{gray}{{\textit{Keyframe Infilling}}}} \\
    \textcolor{gray}{HY-Motion} & \textcolor{gray}{13.9} & \textcolor{gray}{1.03} & \textcolor{gray}{36.82} & \textcolor{gray}{0.206} & \textcolor{gray}{0.319} & \textcolor{gray}{0.394} & \textcolor{gray}{6.618} \\
    CondMDI~\cite{cohan2024flexible} & 2.72 & \textbf{0.92} & 0.803 & 0.160 & 0.277 & 0.366 & 6.471 \\
    MotionLab~\cite{guo2025motionlab} & 4.01 & 1.44 & 0.117 & 0.495 & 0.673 & 0.776 & 3.131 \\
    \cdashline{1-8}
    \textcolor{gray}{Architectures} & \multicolumn{7}{c}{\kern-1.5cm\textcolor{gray}{\textit{\method-Expert}}\kern1.5cm} \\
    ControlNet & 3.98 & 1.19 & 0.635 & 0.472 & 0.671 & 0.778 & 3.166 \\
    AdaLN & 4.15 & 3.18 & 0.102 & \textbf{0.530} & \textbf{0.727} & \textbf{0.830} & \textbf{2.875} \\
    Sequential Concat & 3.09 & 1.04 & 0.655 & 0.496 & 0.691 & 0.793 & 3.024 \\
    Temporal Fusion & \textbf{2.06} & \underline{0.95} & \underline{0.075} & 0.496 & 0.698 & 0.794 & 3.032 \\
    \cdashline{1-8}
    \multicolumn{8}{c}{\textcolor{gray}{\textit{\method-Unified}}} \\
    Temporal Fusion & \underline{2.67} & \underline{0.95} & \textbf{0.040} & \underline{0.514} & \underline{0.707} & \underline{0.804} & \underline{2.974} \\
    \bottomrule
    \end{tabular}%
    }
\end{table}

%% file: Figure/qual_t2m.tex
\begin{figure*}[p]
    \centering
    \includegraphics[width=\textwidth]{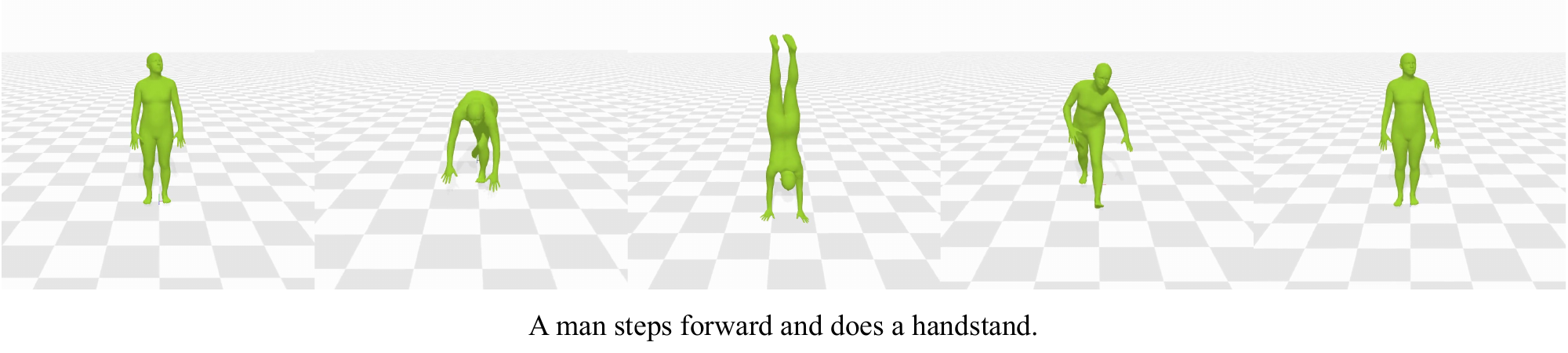}
    \vspace{-0.7cm}
    \caption{Qualitative result on Text-to-Motion Generation.}
    \label{fig:qual_t2m}
\end{figure*}

%% file: Figure/qual_prediction.tex
\begin{figure*}[p]
    \centering
    \includegraphics[width=\textwidth]{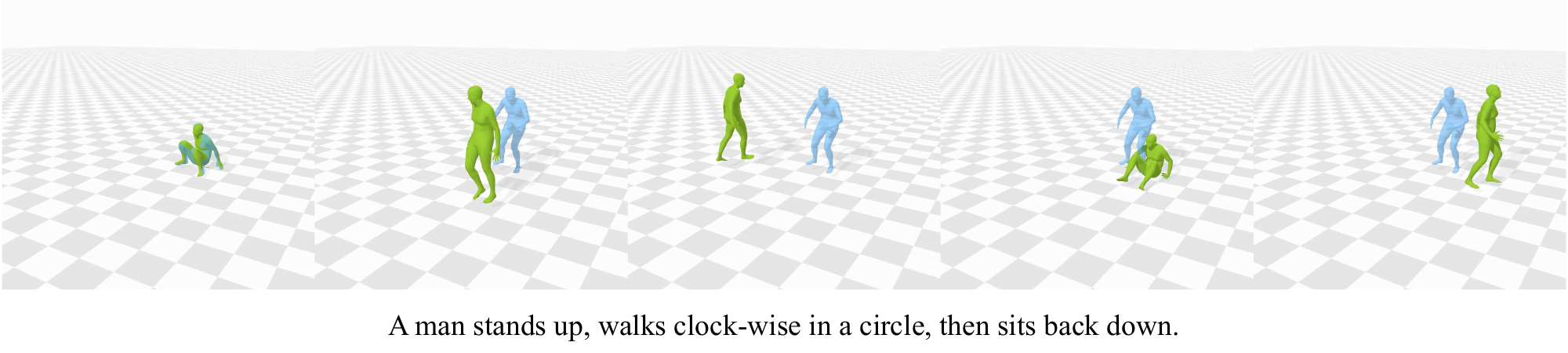}
    \vspace{-0.7cm}
    \caption{Qualitative result on Temporal Inpainting (Prediction). Blue frames denote known keyframes.}
    \label{fig:qual_prediction}
\end{figure*}

%% file: Figure/qual_backcasting.tex
\begin{figure*}[p]
    \centering
    \includegraphics[width=\textwidth]{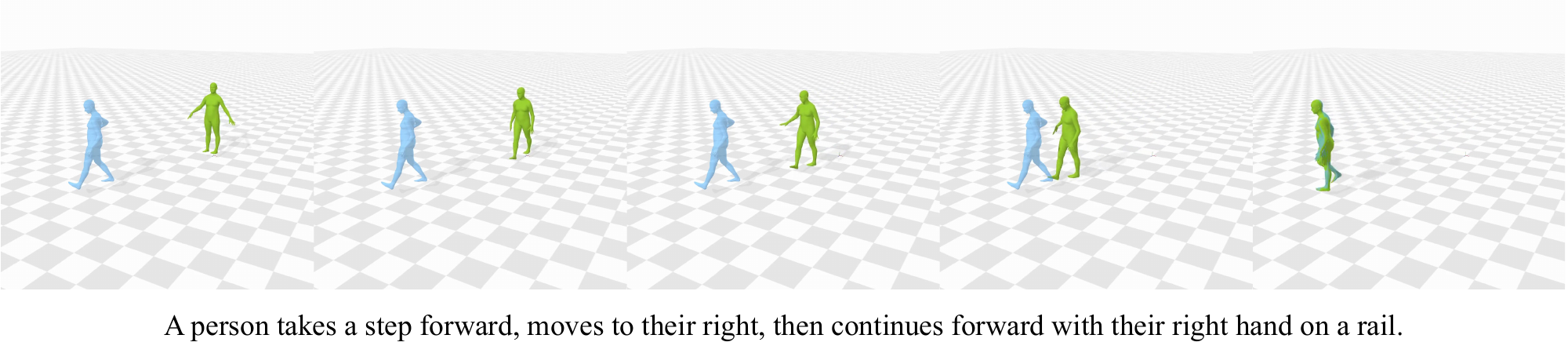}
    \vspace{-0.7cm}
    \caption{Qualitative result on Temporal Inpainting (Backcasting). Blue frames denote known keyframes.}
    \label{fig:qual_backcasting}
\end{figure*}

%% file: Figure/qual_inbetween.tex
\begin{figure*}[p]
    \centering
    \includegraphics[width=\textwidth]{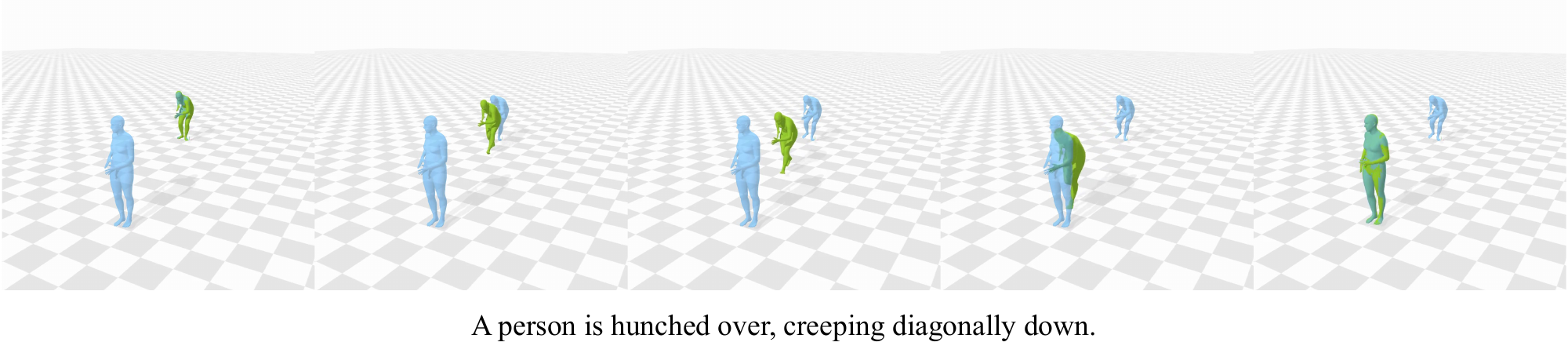}
    \vspace{-0.7cm}
    \caption{Qualitative result on Temporal Inpainting (In-betweening). Blue frames denote known keyframes.}
    \label{fig:qual_inbetween}
\end{figure*}

%% file: Figure/qual_infilling.tex
\begin{figure*}[p]
    \centering
    \includegraphics[width=\textwidth]{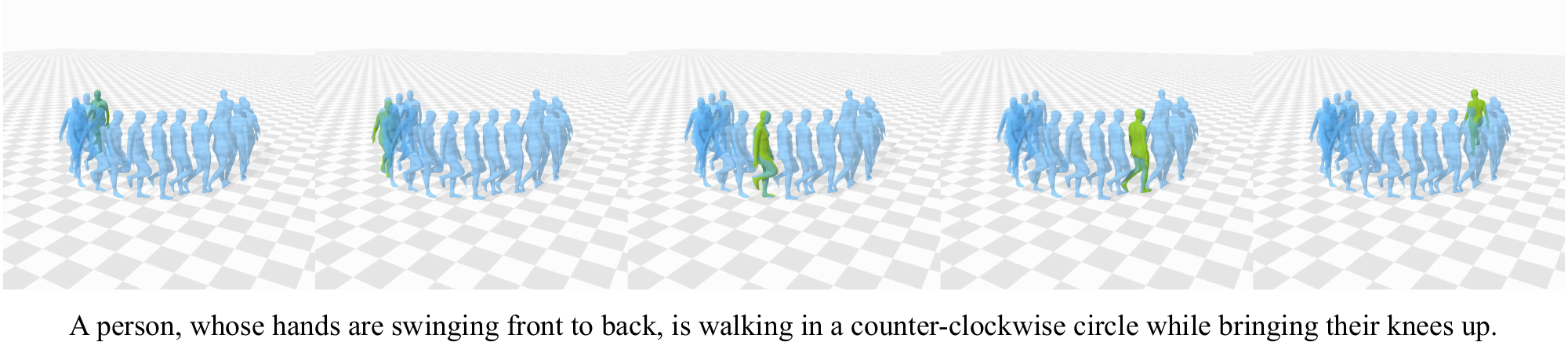}
    \vspace{-0.7cm}
    \caption{Qualitative result on Temporal Inpainting (Keyframe Infilling). Blue frames denote known keyframes.}
    \label{fig:qual_infilling}
\end{figure*}

%% file: Figure/qual_motionediting.tex
\begin{figure*}[p]
    \centering
    \includegraphics[width=\textwidth]{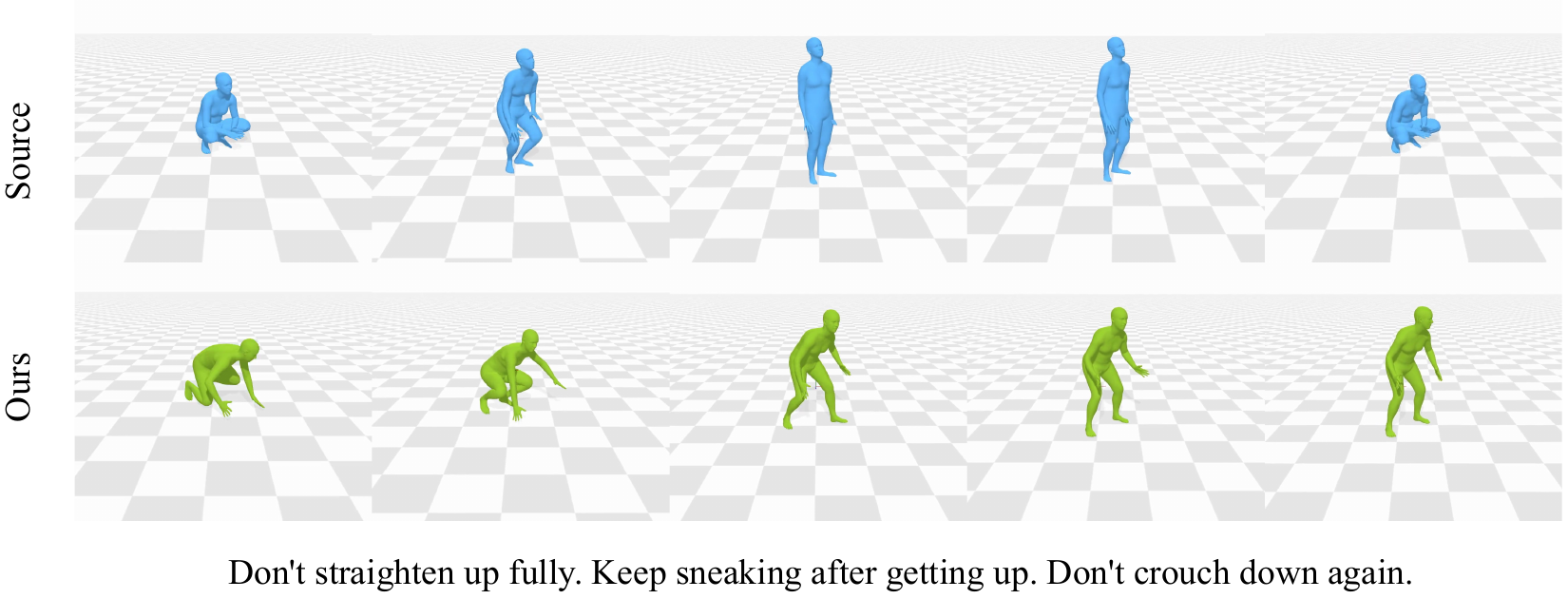}
    \vspace{-0.7cm}
    \caption{Qualitative result on Instruction-Based Motion Editing.}
    \label{fig:qual_motionediting}
\end{figure*}

%% file: Figure/qual_traj.tex
\begin{figure*}[p]
    \centering
    \includegraphics[width=\textwidth]{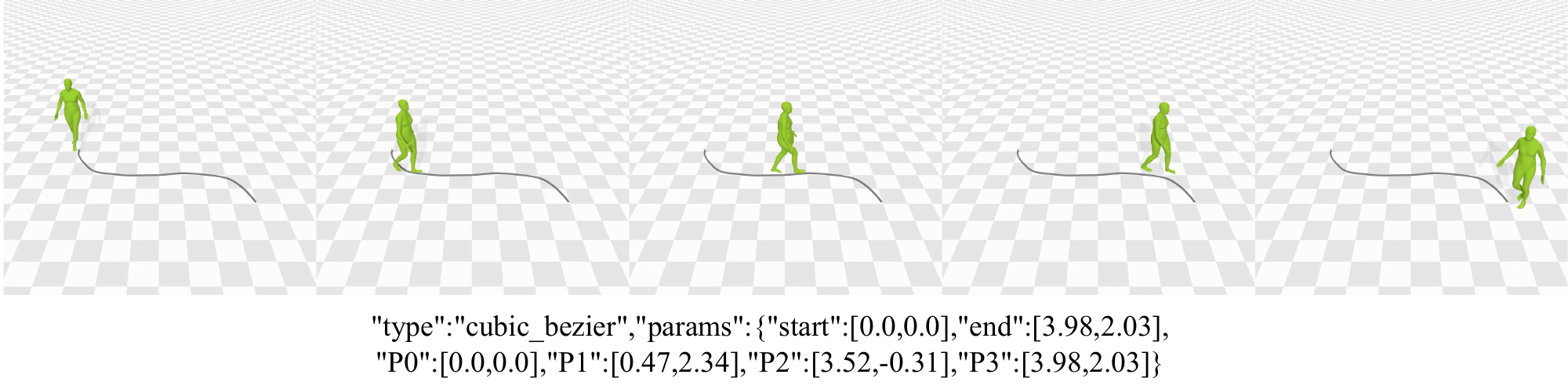}
    \vspace{-0.7cm}
    \caption{Qualitative result on Trajectory Following.}
    \label{fig:qual_traj}
\end{figure*}

%% file: Figure/qual_obstacle.tex
\begin{figure*}[p]
    \centering
    \includegraphics[width=\textwidth]{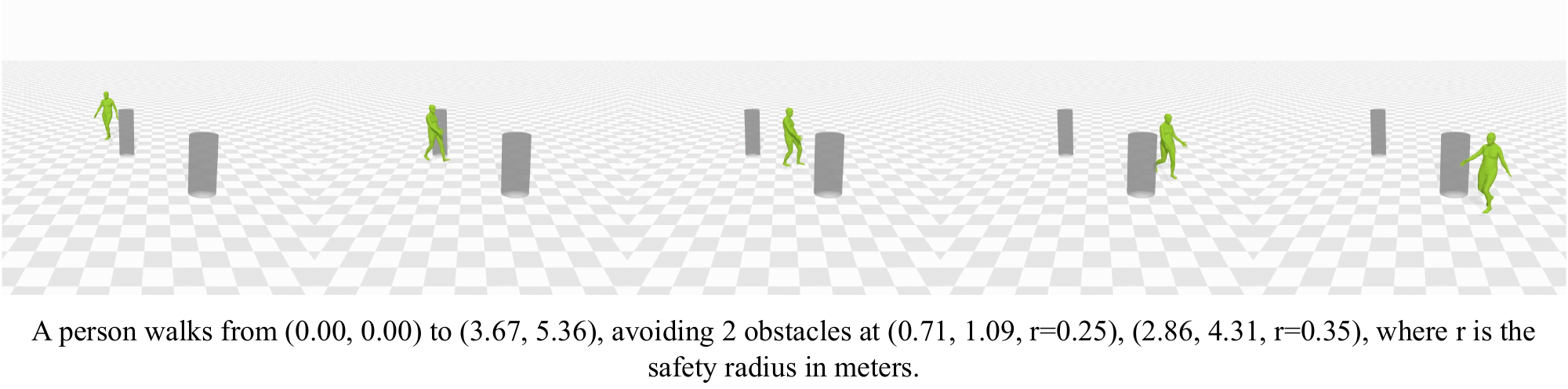}
    \vspace{-0.7cm}
    \caption{Qualitative result on Obstacle Avoidance.}
    \label{fig:qual_obstacle}
\end{figure*}

%% file: Figure/qual_react.tex
\begin{figure*}[p]
    \centering
    \includegraphics[width=\textwidth]{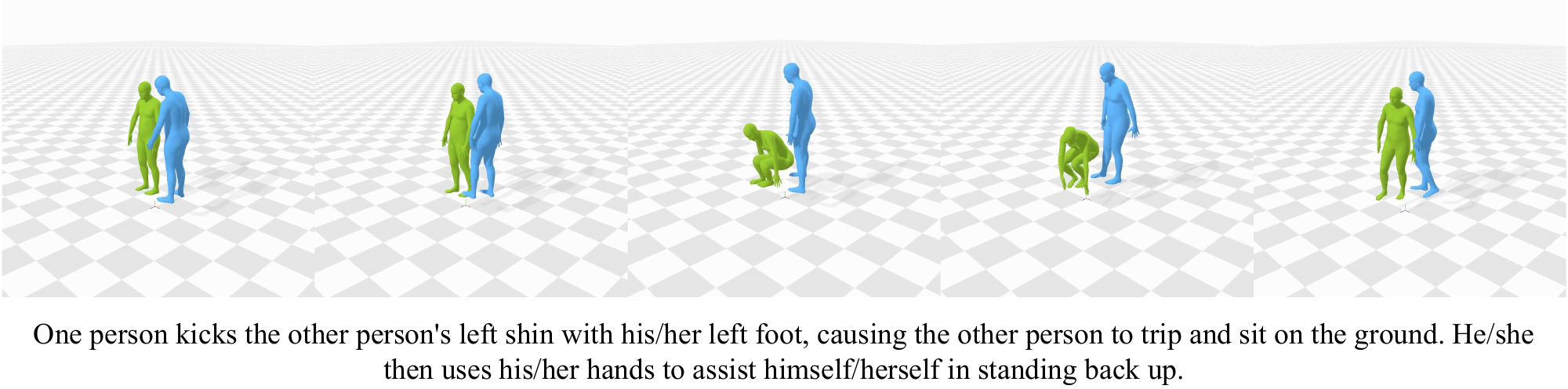}
    \vspace{-0.7cm}
    \caption{Qualitative result on Reaction Generation.}
    \label{fig:qual_react}
\end{figure*}